%% file: root.tex
\documentclass[letterpaper, 10 pt, journal, twoside]{IEEEtran}

\usepackage{graphics} 
\usepackage{epsfig} 
\usepackage{times} 
\usepackage{amsmath} 
\usepackage{amssymb}  

\usepackage{bm}
\usepackage{graphicx}
\usepackage{color}
\usepackage{import}
\usepackage[font=footnotesize]{caption} 
\usepackage{subcaption}
\usepackage{hyperref}
\usepackage{cite}
\usepackage{float}
\usepackage[percent]{overpic} 
\usepackage{algorithm2e}
\usepackage{gensymb} 
\usepackage[para,online,flushleft]{threeparttable} 
\usepackage{array, makecell}
\usepackage{tabularx} 
\usepackage{xcolor} 
\usepackage[percent]{overpic} 
\usepackage{moresize} 

\def\BibTeX{{\rm B\kern-.05em{\sc i\kern-.025em b}\kern-.08em
    T\kern-.1667em\lower.7ex\hbox{E}\kern-.125emX}}
\newcommand{\algName}{ShapeICP}
\newcommand{\supplementary}{the Appendix}
\newcommand{\shapec}{\mathbf{c}} 
\newcommand{\inlinetitle}{\bf} 
\newcommand{\interpwt}{\mu} 
\newcommand{\diffcolor}[1]{\textcolor{black}{#1}} 

\begin{document}

\title{\algName: Iterative Category-Level Object Pose and Shape Estimation from Depth}

\author{Yihao Zhang$^{1}$, Harpreet S. Sawhney$^{2}$, John J. Leonard$^{1}$
\thanks{Manuscript received: June 11, 2025; Revised: August 22, 2025; Accepted: September 19, 2025.}
\thanks{This paper was recommended for publication by Editor Sven Behnke upon evaluation of the Associate Editor and Reviewers’ comments.}
\thanks{$^{1}$Yihao Zhang and John J. Leonard are with the Computer Science and Artificial Intelligence Laboratory, Massachusetts Institute of Technology {\tt\footnotesize \{yihaozh, jleonard\}@mit.edu}. $^{2}$Harpreet S. Sawhney is with Amazon Robotics {\tt\footnotesize hasawhne@amazon.com}.}
\thanks{Digital Object Identifier (DOI): see top of this page.}
\thanks{Code available at: \url{https://github.com/bluekaze/ShapeICP}}}

\markboth{IEEE Robotics and Automation Letters. Preprint Version. Accepted September, 2025}
{Zhang \MakeLowercase{\textit{et al.}}: \algName: Category-Level Object Pose and Shape Estimation} 


\maketitle

\begin{abstract}
Category-level object pose and shape estimation from a single depth image has recently drawn research attention due to its potential utility for tasks such as robotics manipulation. The task is particularly challenging because the three unknowns, object pose, object shape, and model-to-measurement correspondences, are compounded together, but only a single view of depth measurements is provided. Most of the prior work heavily relies on data-driven approaches to obtain solutions to at least one of the unknowns, and typically two, \diffcolor{risking generalization failures if not designed and trained carefully}. The shape representations used in the prior work also mainly focus on point clouds and signed distance fields (SDFs). In stark contrast to the prior work, we approach the problem using an iterative estimation method that does not require learning from pose-annotated data. Moreover, we construct and adopt a novel mesh-based object active shape model (ASM), \diffcolor{which additionally maintains vertex connectivity compared to the commonly used point-based object ASM}. Our algorithm, \algName, is based on the iterative closest point (ICP) algorithm but is equipped with additional features for the category-level pose and shape estimation task. Although not using pose-annotated data, \algName~surpasses many data-driven approaches that rely on pose data for training, opening up a new solution space for researchers to consider.
\end{abstract}


\section{Introduction}
\IEEEPARstart{W}{e} investigate the task of estimating the pose and shape of an object given a single depth image of it. The object instance is only known up to its semantic category. Its exact geometry (i.e. CAD model) is unavailable to us. This problem is common in many scenarios, from warehouse robots to home robots. For example, home robots may require estimating the poses and shapes of tabletop objects to manipulate them. Because of the significance of this task, it has drawn substantial research attention in recent years.


The category-level object pose and shape estimation problem is typically set up in a way that the instance segmentation and semantic classification of the object are performed on the accompanying RGB image with an off-the-shelf method \cite{tian2020shape, chen2021sgpa, wang2021category, chen2020learning, deng2022icaps, akizuki2021asm, lin2021dualposenet, lin2022sar, chen2020category, wang2020directshape, lee2021category, shan2021ellipsdf}. However, even after factoring out the object segmentation problem, inferring the object pose and shape from the segmented depth image remains challenging due to the three entangled unknowns: object pose, object shape, and model-to-measurement correspondences. It requires us to know two of the three unknowns to reduce the problem to an easily solvable one. For instance, if we know the object shape and how the measurements correspond to the points on the shape, it is reduced to the problem of aligning two point sets, which the Umeyama algorithm \cite{umeyama1991least} can solve. If the object pose and the correspondences between the measurements and a shape template are known, one can transform and deform the template to fit the measurements to obtain the shape. Researchers often recognize the correspondences also as unknowns and deform the template \cite{hanocka2020point2mesh} using the Chamfer loss, where the closest points are the correspondences \cite{gkioxari2019mesh, wang2018pixel2mesh, hanocka2020point2mesh}. However, with the three variables all unknown, the problem is extremely difficult.

Most prior methods exploit neural networks to learn a mapping from inputs such as the RGB-D image and a categorical shape prior to one or two of the three unknowns. Although these learning methods greatly factor away the complexities in the problem by leveraging data and have achieved excellent performance on benchmarks, they face potential failures caused by domain changes and the tedious data curation process required for training when they are deployed to a new environment. The ground truth pose of an object is especially painstaking to annotate since so far there has been no scalable and easily accessible approach to measure the pose of an object \diffcolor{in the real world}. On the other hand, virtually all the prior methods focus on either point clouds or signed distance fields (SDFs) as the shape representation. Point clouds are flexible but lack the notion of surface, while SDFs entail unfavorable computational complexity, which often scales cubically with resolution.

In this work, we take a radically different approach \diffcolor{and make the following contributions}. First, in contrast to prior methods, we adopt a novel mesh-based active shape model (ASM) \cite{cootes1995active} (instead of the commonly used point-based or SDF-based object ASM) as the shape representation. Meshes are an \diffcolor{explicit} geometric representation, unlike SDFs or occupancy grids where the geometry is represented implicitly and some processing, such as finding the zero level set (SDF) or marching cubes \cite{lorensen1998marching} (occupancy grid), is required to extract the geometry. \diffcolor{A mesh-based ASM also incorporates the notion of surface (or vertex connectivity), being superior to point ASM. However, the addition of the connectivity information requires a different ASM construction methodology, which is also our contribution}.

Second, although methods such as self-supervised learning and sim-to-real transfer can help circumvent the annotation difficulty, we seek a different avenue by devising a \diffcolor{learning-free iterative estimation algorithm based on the iterative closest point (ICP) algorithm \cite{besl1992method} and alternating minimization. Third, we further adapt a set of robust estimation techniques, such as expectation-maximization, multi-hypothesis tracking, and shape classification, to cope with the local minimum problem in the estimation process. The option of using an image-based shape classification network for shape initialization only requires training on the ground-truth object shape, which is practically more accessible than the ground-truth pose since object scanning and depth fusion techniques \cite{hornung2013octomap} are mature.}

Lastly, we evaluate our method on the NOCS REAL dataset \cite{wang2019normalized} and find it better or comparable to many learning-based methods, even though our method is at the disadvantage that no pose-annotated data are used. The evaluation of our method also reveals what can already be accomplished without learning and why learning may be needed in some places. These insights are traditionally difficult to draw from the evaluation of black-box style learning-based methods. \diffcolor{We provide more extensive evaluation and analysis in \supplementary.}

\section{Related Work} \label{sec: relwork}
We will review the related work in terms of which of the three unknowns, object pose, object shape, and model-to-measurement correspondences, are predicted by neural networks. We will also discuss shape representations, particularly ASM, used in the literature.

\subsection{Category-Level Pose and Shape Estimation} \label{sec: relest}
{\inlinetitle Predicting shape and correspondences.} The seminal work NOCS \cite{wang2019normalized} belongs to this category. The predicted NOCS map encodes the depth profile (i.e. shape) of the observed object in its canonical pose. The NOCS map also corresponds pixel-to-pixel to the measured depth image. Another work is \cite{lee2021category} where the authors first use Mesh R-CNN \cite{gkioxari2019mesh} to predict a metric-scale mesh of the object shape from the image and then predict the NOCS map to solve for the pose. A notable line of work \cite{tian2020shape, chen2021sgpa, wang2021category} attempts to predict the deformation applied to a categorical shape prior to obtain the object shape and a correspondence matrix to associate the model (i.e. the deformed prior) points to the measured depth points. For all these methods, the final step is the Umeyama algorithm \cite{umeyama1991least} to solve for the object pose.

{\inlinetitle Predicting shape and pose.} CASS \cite{chen2020learning} first learns a point cloud based shape embedding space. During inference, the shape latent code is predicted given the RGB-D image patch. The predicted latent code is both decoded to the shape and fed to a module for pose prediction. CenterSnap \cite{irshad2022centersnap} and ShAPO \cite{irshad2022shapo} build a YOLO-style \cite{redmon2016you} one-shot estimation pipeline. iCaps \cite{deng2022icaps} selects the rotation by comparing the depth image features to the features in a pre-computed code book and predicts the DeepSDF \cite{park2019deepsdf} latent vector as the shape representation. Other methods that directly regress shape and pose to various extents include \cite{akizuki2021asm, lin2021dualposenet, bruns2022sdfest}.

{\inlinetitle Predicting pose.} This category of methods bypasses shape estimation and estimates only the pose. FS-Net \cite{chen2021fs} extracts rotation-aware features through a scale- and shift-invariant 3D graph convolution network. It decouples the rotation into two perpendicular vectors for prediction. SAR-Net \cite{lin2022sar} trains neural networks to transform a categorical template to the same rotation as the observed point cloud and complete the observed point cloud by mirroring about the symmetry plane. These network outputs are used to compute the pose.

{\inlinetitle Predicting correspondences.} \cite{shi2023optimal} assumes known correspondences provided by neural keypoint matching and solves an alignment problem between the model keypoints and the measurement keypoints. Instead of only estimating the $\mathrm{SIM}(3)$ pose as in Umeyama \cite{umeyama1991least}, the method additionally estimates the parameters for a point-based active shape model (ASM) \cite{cootes1995active} of the object. The ASM is linear and can be treated similarly to translation. The method also features a graph-theoretic approach and an outlier-robust solver to handle false correspondences. \cite{pavlakos20176} is an earlier work that essentially shares the same problem formulation as \cite{shi2023optimal} but with a different solver that does not consider the outlier issue.

{\inlinetitle Optimization-based.} Neural Analysis-by-Synthesis \cite{chen2020category} leverages neural rendering to render an image from the object shape code and pose. The rendered image is compared with the actual image to optimize the shape code and pose. DirectShape \cite{wang2020directshape} sets up the optimization losses by silhouette matching and left-right stereo photo consistency using an SDF-based ASM. Strong regularization, such as the ground plane constraint, is engaged to alleviate the optimization difficulty. ELLIPSDF \cite{shan2021ellipsdf} employs a bi-level shape representation to coarsely fit the object segmentation for initialization and then finely optimize the pose and shape using multiple views.

{\inlinetitle Summary.} As we have seen, previous methods commonly adopt data-driven solutions. Pose-annotated data are exploited in almost all the prior work, either explicitly or implicitly. Even the optimization-based methods may actually require extra information such as multiple views \cite{shan2021ellipsdf} or application-specific constraints \cite{wang2020directshape}. Our method works completely without pose-annotated data. The addition of an optional shape classification network merely requires shape-annotated data, which also falls into the rare category of predicting only the shape, a direction not well explored by prior work.

\subsection{Shape Representation -- Active Shape Model}
The active shape model (ASM) \cite{cootes1995active} has been employed in the category-level object pose and shape estimation task \cite{wang2020directshape, pavlakos20176, shi2023optimal, akizuki2021asm}, and also in the 3D object shape reconstruction task \cite{kundu20183d}. The ASM has the advantage of being a simple linear model that is more efficient to back-propagate through during optimization compared to a neural network. The object ASM is primarily point-based \cite{pavlakos20176, shi2023optimal, akizuki2021asm} and SDF-based \cite{wang2020directshape, kundu20183d}. It is interesting to explore the mesh-based ASM further.

\section{Method}
\subsection{Overview}
Our task is to estimate the $\mathrm{SIM}(3)$ pose and the shape of an object which is only known up to its semantic category given a single depth image of it. We assume that the instance segmentation of the object and its class label are provided by an off-the-shelf method such as Mask RCNN \cite{vuola2019mask}, as is typically done in the prior work \cite{chen2020category, wang2020directshape, tian2020shape, chen2021sgpa, wang2021category, chen2020learning, akizuki2021asm, deng2022icaps, lin2021dualposenet, lin2022sar}. We adopt a mesh-based object active shape model (ASM) \cite{cootes1995active} for its flexibility and surface representability. After obtaining the ASM offline (Section \ref{sec: asm}), we estimate the pose and shape by transforming and fitting the shape model to the observed depth points. The optimization is realized by an augmented iterative closest point (ICP) algorithm which, in addition to the existing pose estimation step, has a shape deformation step for shape estimation (Section \ref{sec: altalg}). We thus name our algorithm \algName. However, a significant challenge is that ICP is a local solver, and so is the base form of \algName. To overcome this, \algName~uses multi-hypothesis tracking for rotation estimation, expectation maximization (EM) for robust correspondence handling, and shape classification for shape initialization (Section \ref{sec: copemin}). The full pseudo-algorithm of \algName~can be found in Section \ref{sec: pseudoalg}. 

\subsection{Mesh-based Object Active Shape Model} \label{sec: asm}
\begin{figure}[h]
\centering
\begin{overpic}[width=1.0\linewidth, tics=5]{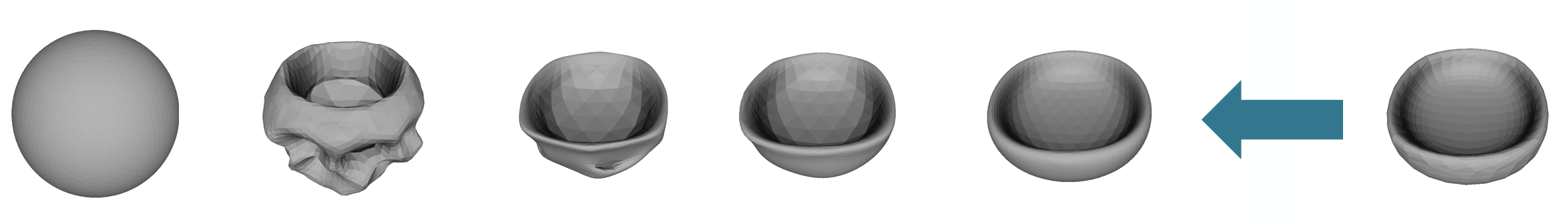} 
    \put(0,0){\diffcolor{\ssmall{initial template}}}
    \put(60,0){\diffcolor{\ssmall{deformed template}}}
    \put(88.2,0){\diffcolor{{\ssmall{target mesh}}}}
\end{overpic}
\caption{An example of the template deformation process.}
\label{fig: deform}
\end{figure}

{\inlinetitle Template deformation.} Mesh-based ASMs have been applied to human faces \cite{blanz2023morphable} but not yet in the task of object pose and shape estimation. The bottleneck is that object models from a database have different numbers of vertices and different vertex connectivity even within the same category, making it difficult to draw corresponding vertices \diffcolor{\cite{akizuki2021asm} and maintain consistent vertex connectivity across models for PCA (elaborated in \supplementary). To solve these challenges}, we deform a template mesh to wrap around each object model and run PCA on the vertices of the deformed templates for all the object models within a category. Specifically, we denote a mesh by $\mathcal{M = (V, E, F)}$, where $\mathcal{V} = \{v_i\}_{i=1}^V$ are the $V$ vertices in the mesh, $\mathcal{E} = \{e_i\}_{i=1}^E$ is the set of $E$ tuples of the indices of a pair of vertices connected by an edge, $\mathcal{F} = \{f_i\}_{i=1}^F$ is the set of $F$ tuples of indices of vertices on the same face. $\mathcal{E}$ or $\mathcal{F}$ defines the vertex connectivity. To deform the template (source) mesh $\mathcal{M}_s$ to a target mesh $\mathcal{M}_t$, we randomly sample points on the mesh surface with $N$ points from the template mesh $p_n \sim \mathcal{M}_s$ and $M$ points from the target mesh $q_m \sim \mathcal{M}_t$ (here $N = M$). \diffcolor{As in \cite{wang2018pixel2mesh, gkioxari2019mesh, ravi2020pytorch3d},} the deformation can be done by minimizing losses:
\begin{equation}
    \mathcal{L}_c = \frac{1}{N}\sum_{n}{\min_{m}{\|p_n - q_m\|_2^2}} + \frac{1}{M}\sum_{m}{\min_{n}{\|p_n - q_m\|_2^2}}
\label{eq: chamfer}
\end{equation}
which is the Chamfer distance between the two sets of sampled points. To regularize the template mesh during the deformation process, a normal consistency loss is imposed on the template mesh:
\begin{equation}
    \mathcal{L}_n = \frac{1}{E}\sum_{e}{1 - \cos{(\mathbf{n}_{f_{e+}}, \mathbf{n}_{f_{e-}})}}
\label{eq: normal}
\end{equation}
where $\mathbf{n}_{f_{e+}}$ and $\mathbf{n}_{f_{e-}}$ are the normals of the two faces that share edge $e$, and $\cos$ is the cosine distance. This loss regularizes the surface normals to be smooth. To avoid overly long outlier edges in the template mesh, an edge length loss is added:
\begin{equation}
    \mathcal{L}_e = \frac{1}{E}\sum_{e}{\|v_{e+} - v_{e-}\|_2^2}
\label{eq: edge}
\end{equation}
where $v_{e+}$ and $v_{e-}$ are the vertices connected by the edge $e$. A Laplacian smoothing loss is also enforced on the template mesh to encourage the vertices to move along with their neighbors and potentially avoid mesh self-intersection:
\begin{equation}
    \mathcal{L}_l = \frac{1}{V}\sum_{i}{\|v_i - \frac{1}{|\mathcal{N}(i)|}\sum_{j\in\mathcal{N}(i)}{v_j}\|_2}
\label{eq: lap}
\end{equation}
where $\mathcal{N}(i) = \{j\}~\forall(i,j)\in\mathcal{E}$ is the set of neighboring vertices of vertex $v_i$. The overall objective function is $\min_{\mathcal{V}_s}\mathcal{L}_c + \lambda_n\mathcal{L}_n + \lambda_e\mathcal{L}_e + \lambda_l\mathcal{L}_l$, where the $\lambda$'s are the weights for the loss terms. This optimization is carried out by stochastic gradient descent (SGD) \cite{robbins1951stochastic} starting from a spherical template with $p_n$ and $q_m$ resampled in each step. An example of the process is visualized in Fig. \ref{fig: deform}.
\setlength{\intextsep}{2.5pt}

\begin{figure}[h]
\centering
\includegraphics[width=0.9\linewidth]{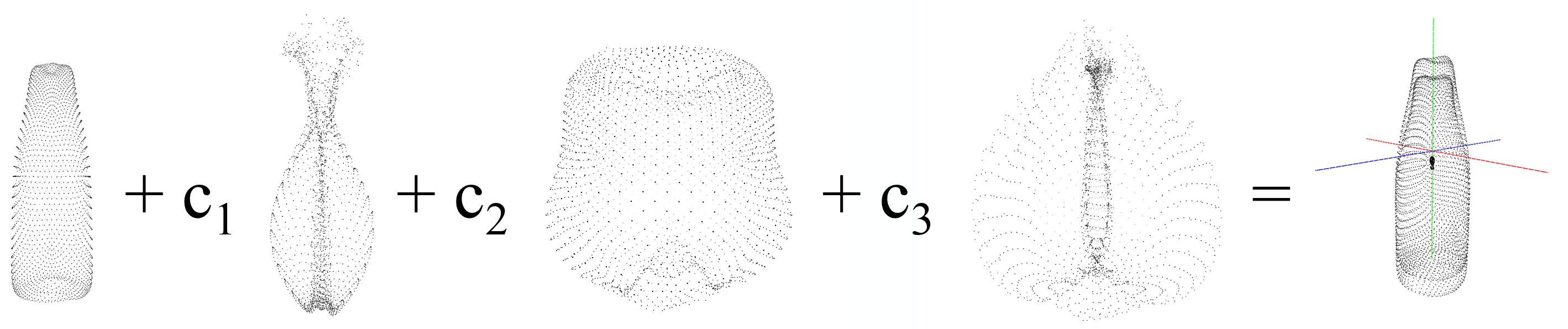}
\caption{Visualization of the vertices of the bases and the final shape in an example ASM. The final shape on the right shows how the mean shape is displaced by the weighted bases (the scale of the bases is much smaller than the mean, so the bases on the left are enlarged for clarity).}
\label{fig: asmviz}
\end{figure}

{\inlinetitle PCA.} Now that we have acquired a set of deformed templates that take on the model shapes across a category and have a consistent number of vertices and the same connectivity, we can build a feature vector for each model by concatenating the vertices of its deformed template, i.e. $[v_1...v_{V_s}]$, and then running PCA on all the feature vectors to obtain our category-level mesh-based ASM. A vertex in the final ASM with $K$ principal components is expressed as
\begin{equation}
    v_i = b_{0,i} + \sum_{k=1}^{K}{c_k b_{k,i}}
\label{eq: asm}
\end{equation}
where $b_{0,i}$ is the vertex in the mean shape corresponding to $v_i$, $b_{k,i}$ is the corresponding vertex in the basis $k$, and $\{c_k\}_{k=1}^K$ are the weights for the bases. These weights parameterize the shape. Fig. \ref{fig: asmviz} visualizes the vertices of the bases and the final shape in an example ASM. The edges ($\mathcal{E}$) and faces ($\mathcal{F}$) (i.e. connectivity) of the mesh-based ASM are inherited from the template mesh.

\subsection{Alternating Pose and Shape Optimization} \label{sec: altalg}
{\inlinetitle Formulation.} To introduce our \algName~formulation, we need sample points on the ASM. Each sampled point can be expressed as
\begin{equation}
    p_n = \sum_{i\in\mathsf{F}(n)}{\interpwt_{n,i}v_i}~~\text{s.t.} \sum_{i\in\mathsf{F}(n)}{\interpwt_{n,i}} = 1
\label{eq: sample}
\end{equation}
where $\mathsf{F}(n)$ is a function that maps the sample index $n$ to a face randomly selected with probability proportional to the face area, and $\interpwt_{n,i}$ are randomly generated interpolation weights for this sample. These interpolation weights are fixed during the later optimization. Combining with \eqref{eq: asm}, $p_n = \sum_{i\in\mathsf{F}(n)}{\interpwt_{n,i}(b_{0,i} + \sum_{k}{c_k b_{k,i}})}$. We now introduce the \algName~formulation:
\begin{equation}
    \min_{\substack{R\in\mathrm{SO}(3), t\in\mathbb{R}^3, s\in\mathbb{R}_{++} \\ \shapec = [c_1...c_K]\in\mathbb{R}^K}}{\frac{1}{M}\sum_{m=1}^{M}{\min_{n}{\|sRp_n + t - q_m\|_2^2}}}
\label{eq: shapeicp}
\end{equation}
where $\{q_m\}_{m=1}^M$ (overloaded notation) are now the object depth points segmented from the depth image and back-projected to 3D using the camera intrinsics, $R$, $t$, and $s$ are the object rotation, translation, and scale to be estimated, $\shapec$ which enters the objective function through $p_n$ is the ASM shape code also to be estimated. For each measurement $q_m$, \eqref{eq: shapeicp} encourages the transformed model points to be close to the measurements, which is similar to the ICP objective function \cite{besl1992method} but with the addition of the shape parameter $\shapec$. We also remark that formulation \eqref{eq: shapeicp} equivalently works for a point-based ASM \cite{akizuki2021asm} where $p_n$ is simply a point in the ASM point cloud. However, a mesh-based ASM has the notion of surface and thus is a better geometric representation. Furthermore, a mesh model enables the possibility of rendering to be used in Section \ref{sec: copemin}.

{\inlinetitle Core algorithm.} We make two observations of \eqref{eq: shapeicp}. First, if the shape code $\shapec$ is given, the problem shares the same objective function as ICP. Second, if the $\mathrm{SIM}(3)$ pose is given, the problem mimics the mesh deformation problem introduced in Section \ref{sec: asm} except that \eqref{eq: shapeicp} is only the one-sided Chamfer distance and the optimization variable is the shape code instead of the vertices. These two observations lead to an intuitive alternating minimization algorithm to solve \eqref{eq: shapeicp}:
\begin{itemize}
    \item {\it Pose step:} Associate each measurement $q_m$ to the closest transformed model point $\hat{s}\hat{R}\hat{p}_n + \hat{t}$ (i.e. solving $\min_n$ in \eqref{eq: shapeicp}) using the last estimates $\hat{\shapec}$, $\hat{R}$, $\hat{t}$, and $\hat{s}$. Solve for the incremental $R$, $t$, and $s$ with Umeyama \cite{umeyama1991least} to further align the model points with the measurements while keeping $\shapec$ fixed.
    \item {\it Shape step:} Re-associate each $q_m$ to a model point using the latest estimates. Solve for $\shapec$ as if it is a mesh deformation problem while keeping $R$, $t$, and $s$ fixed.
\end{itemize}
After the pose step, the incremental $R$, $t$, and $s$ are accumulated onto the estimates $\hat{R}$, $\hat{t}$, and $\hat{s}$. Every time $\shapec$ is updated, the model points ($p_n$'s) are re-sampled to ensure sample uniformity as the areas of the mesh faces can change.

{\inlinetitle Shape step.} In the shape step, we include the regularization losses \eqref{eq: normal} -- \eqref{eq: lap}. The final objective function $\mathcal{L}_{sc}$ ($sc$: shape code) in the shape step is $\min_{\shapec}{\mathcal{L}_{sc}} = \min_{\shapec}\mathcal{L}_{ps} + \lambda_n\mathcal{L}_n + \lambda_e\mathcal{L}_e + \lambda_l\mathcal{L}_l$, where $\mathcal{L}_{ps}$ ($ps$: pose shape) refers to the objective function in \eqref{eq: shapeicp} and the $\lambda$'s are the weights. These weights can be reused from Section \ref{sec: asm} if we center and normalize the observed point cloud, and shift and normalize the transformed model points by the same amount (because the ShapeNet \cite{chang2015shapenet} objects in Section \ref{sec: asm} are all centered and normalized). Since this is an iterative coordinate-descent style algorithm, we prefer small increments on $\shapec$. Only one step of gradient descent is performed on $\shapec$ during each shape step. However, we repeat the shape step several times in each iteration. The shape step is skipped in the last few iterations for better empirical performance.


\subsection{Coping with Local Minima} \label{sec: copemin}
\begin{figure}[h]
\centering
\vspace{0.8em} 
\includegraphics[width=0.8\linewidth]{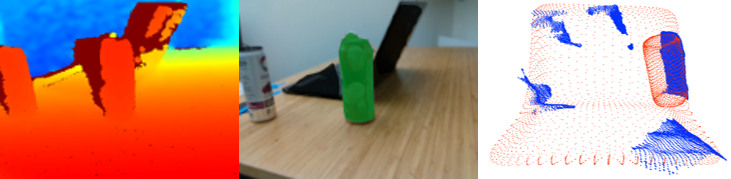}
\caption{\diffcolor{The back-projected depths (blue points) are fragmented due to a partial view and occlusion, which causes ambiguity such that the models (red) fit the measurements well but have pose errors (in collision with each other).}}
\label{fig: ambiguity}
\vspace{-0.2em}
\end{figure}

Category-level object pose and shape estimation given only a single-view depth image naturally comes with large ambiguity (Fig. \ref{fig: ambiguity}) since many configurations of the variables may fit equally well to the observed partial object point cloud. In other words, the optimization landscape exhibits many local minima, which hurt the performance of a local solver such as \algName. In this section, we develop several strategies to cope with the local minima.

{\inlinetitle Expectation maximization (EM).} The correspondences between the measured depth points and the model points can be treated as latent variables. Instead of hard one-to-one associations, EM allows softer associations \cite{bishop2006pattern}, reducing the chance of being stuck at a local minimum. Inspired by \cite{parkison2018semantic}, we integrate EM into our \algName~formulation. The new objective function $\mathcal{L}_{em}$ derived from EM is (details about the probability models, independence assumptions, and derivation can be found in \supplementary):
\begin{equation}
    \min_{\substack{R\in\mathrm{SO}(3), t\in\mathbb{R}^3, s\in\mathbb{R}_{++} \\ \shapec = [c_1...c_K]\in\mathbb{R}^K}}{\sum_{m=1}^M{\sum_{n\in\mathcal{N}_Q(q_m)}{w_{mn}\|sRp_n + t - q_m\|_2^2}}}
\label{eq: emshapeicp}
\end{equation}
where
\begin{equation}
    w_{mn} = \frac{\mathsf{N}(\hat{s}\hat{R}\hat{p}_n + \hat{t} - q_m; 0, \sigma_m)}{2\sigma^2_m\sum_{n\in\mathcal{N}_Q(q_m)}{\mathsf{N}(\hat{s}\hat{R}\hat{p}_n + \hat{t} - q_m; 0, \sigma_m)}}
\label{eq: emwt}
\end{equation}
where the hatted quantities $\hat{s}$, $\hat{R}$, $\hat{t}$, and $\hat{p}_n$ ($\hat{p}_n$ as a result of $\hat{\shapec}$) are the estimates from the last step and do not participate in the current optimization, $\mathcal{N}_Q(q_m)$ finds the $Q$ closest transformed model points to $q_m$ (i.e. the top $Q$ solutions to $\arg\min_{n}{\|sRp_n + t - q_m\|_2^2}$), and $\mathsf{N}(\cdot; 0, \sigma_m)$ is a zero-mean Gaussian distribution with isotropic covariance $\Sigma_m = \mathsf{diag}(\sigma_m^2)$. EM yields an intuitive formulation that considers $Q$ likely correspondences weighted by their probabilities computed from the last-step estimates rather than one hard correspondence. This increases the field of view of the model-measurement association step and can help move out of local minima. Both the pose and shape steps can use \eqref{eq: emshapeicp}.

{\inlinetitle Multi-hypothesis estimation.} Initialization is critical in avoiding local minima. The translation can be initialized as the centroid of the observed point cloud $\bar{q} = \frac{1}{M}\sum_m{q_m}$. We find it better to begin with a small initial scale such as the average distance to the center $\frac{1}{M}\sum_m{q_m - \bar{q}}$. Without any neural network, the shape code is initialized to be the mean code for the category $\bar{\shapec} = \frac{1}{U}\sum_u{\tilde{\shapec}_u}$ where $U$ is the total number of database object models for the category. However, the rotation initial guess is not readily available. We thus track multiple rotation hypotheses in parallel and drop unpromising hypotheses quickly to save computation. Specifically, we start with the base $\mathrm{SO}(3)$ grid from \cite{yershova2010generating}, as is also used by \cite{bruns2022sdfest}. The grid has 2304 discrete rotations covering $\mathrm{SO}(3)$. During the course of optimization, we drop hypotheses according to the following score functions:

{\it Objective function.} The mean residual in \eqref{eq: shapeicp} $\mathcal{S}_r = \frac{1}{M}\sum_m{r_m} = \frac{1}{M}\sum_m{\min_{n}{\|\hat{s}\hat{R}\hat{p}_n + \hat{t} - q_m\|_2^2}}$ plus the standard deviation of the residuals \small$\mathcal{S}_\sigma = \mathsf{std}(\{r_m\}_{m=1}^M)$\normalsize is used. We find that adding the standard deviation improves the results, as we hope to have relatively uniform residuals over all the depth measurement points. 

{\it Symmetry check.} Man-made objects often have mirror symmetry (e.g. laptops) and rotational symmetry (e.g. bottles and cans). If the estimated pose and shape are correct, manipulating the observed point cloud with respect to the estimated plane and axis of symmetry should still result in a low residual. Let $\hat{T} = [\hat{s}\hat{R}~\hat{t};~0~0~0~1]$ be the $\mathrm{SIM}(3)$ pose estimate, and $\hat{p}_{h,n} = [\hat{p}_n; 1]$ and $q_{h,m} = [q_m; 1]$ be the homogeneous coordinates. The symmetry score $\mathcal{S}_{\psi,r}$ is
\begin{equation}
    \frac{1}{M}\sum_m{r_{\psi,m}} = \frac{1}{M}\sum_m{\min_n{\|\hat{T}\hat{p}_{h,n} - \hat{T}T_\psi\hat{T}^{-1}q_{h,m}\|_2^2}}
\label{eq: symmscore}
\end{equation}
where $T_\psi = [R_\psi~0;~0~0~0~1]$ is a symmetry operation such as rotation around the axis of symmetry or reflection with respect to the plane of symmetry (in which case $\det(R_\psi) = -1$). \eqref{eq: symmscore} essentially transforms the observed point cloud to the object canonical pose using the current estimate $\hat{T}$, applies $R_\psi$, and transforms back so that the order of magnitude of $\mathcal{S}_{\psi,r}$ is similar to $\mathcal{S}_r$. The standard deviation score is $\mathcal{S}_{\psi,\sigma} = \mathsf{std}(\{r_{\psi,m}\}_{m=1}^M)$ and the total symmetry score is $\mathcal{S}_{\Psi} = \frac{1}{\Psi}\sum_\psi{\mathcal{S}_{\psi,r} + \mathcal{S}_{\psi,\sigma}}$ for a total of $\Psi$ different symmetry operations (multiple rotation symmetry operations of different angles are possible).

{\it Depth rendering.} Taking advantage of the mesh-based representation, we render the current estimates to a depth image and compare with the observed depth image of the object (after applying the segmentation mask). The depth rendering score is
\begin{equation}
    \mathcal{S}_{dr} = \frac{1}{|\Omega(D)|}\sum_{d\in\Omega(D)}{[\mathsf{R}(\hat{T}, \mathcal{M}(\hat\shapec), K_c)_d - D_d]^2}
\label{eq: drender}
\end{equation}
where $D$ is the masked observed depth image, $d\in\Omega(D)$ indexes pixels in the image space, $\mathsf{R}$ is the renderer, $\mathcal{M}(\hat\shapec)$ is the mesh model built from $\hat\shapec$, and $K_c$ includes the known camera intrinsics and other camera parameters (such as the image size). We use the renderer from \cite{ravi2020pytorch3d} and assign the same background value to the rendered image and the masked observed depth image. The depth rendering helps to identify an over-sized model (which may still give low $\mathcal{S}_r$ if a portion of it fits the measurements well) \cite{wang2021dsp}. To avoid computation overflow, we start this depth rendering after the number of hypotheses is low.

The total score to be minimized is $\mathcal{S}_{tot} = \mathcal{S}_r + \mathcal{S}_\sigma + \lambda_\Psi\mathcal{S}_\Psi + \lambda_{dr}\mathcal{S}_{dr}$ where the $\lambda$'s are the weights. When picking top hypotheses, we greedily select the next hypothesis at least some angle ($\theta$) apart from the last chosen hypothesis to alleviate duplicated or very close hypotheses.

{\inlinetitle Shape classification.} To better initialize the shape code $\shapec$, we build a shape classification network whose input is a color-coded normal vector image of the segmented object and the output is the index of the closest database model $u^*$. We choose ResNet-50 \cite{he2016deep} for the network. The normal vector image is computed from the observed depth image and converted from $x, y, z \in [0,1]$ (assuming a unit vector) to RGB values. Only the image patch that contains the object (i.e. the 2D bounding box) is resized to a constant size and fed to the network. We find the ground-truth closest model by computing the Chamfer distance \eqref{eq: chamfer} between the ground-truth shape and all the models in the database (ShapeNet \cite{chang2015shapenet}). The final initial guess is $\tilde{\shapec}_{u^*}$ for the shape code. Note that if a model in the database is not matched to any ground-truth shape, it does not appear as a class in the network output. Therefore, the number of classes is smaller than or equal to the number of ground-truth shapes. 

\subsection{Summary} \label{sec: pseudoalg}
The full estimation algorithm is given in Alg. \ref{alg: shapeicp}.
\input{includes/fullalg}

\section{Experiments}
\subsection{Experiment Setup}
{\inlinetitle Implementation.} We implement our \algName~method with PyTorch \cite{paszke2019pytorch} and PyTorch3D \cite{ravi2020pytorch3d}. For the segmented depth image of an object, we perform statistical outlier removal on the back-projected point cloud as in \cite{akizuki2021asm, manhardt2020cps++}. Moreover, we discard object detections that have fewer than a threshold number of points. All the hyperparameter values are given in \supplementary.

{\inlinetitle Datasets.} ShapeNetCore \cite{chang2015shapenet} is used to compute the ASM for the six categories in the NOCS REAL data \cite{wang2019normalized}. For each category, we exclude weird-looking shapes (e.g. cups with saucers), which is similarly done in \cite{akizuki2021asm}. As our model does not train on any pose-annotated data, synthetic benchmarks are irrelevant. We thus evaluate our method on the NOCS REAL test set. For the shape classification network, we use the NOCS REAL train set for training.

{\inlinetitle Baselines.} We benchmark our \algName~method against several interesting baselines. {\it NOCS} \cite{wang2019normalized} is the seminal work in this research area. {\it Neural Analysis-by-Synthesis} \cite{chen2020category} is an optimization-based method that is not learning-based \diffcolor{for the purpose of pose inference} and does not rely on image sequences or strong application-specific constraints (see Section \ref{sec: relest}). {\it Metric Scale} \cite{lee2021category} is a rare case of adopting a mesh-based shape representation. {\it ASM-Net} \cite{akizuki2021asm} also uses ASM, but it is point-based. Unlike our method, its inference is mainly performed by neural network regression. {\it Shape Prior} \cite{tian2020shape} also estimates the shape by deforming a prior, but the method is learning-based \diffcolor{(i.e. regression-based)}. {\it CenterSnap} \cite{irshad2022centersnap} is a one-shot end-to-end learning-based solution, which is the opposite of our work.

{\inlinetitle  Evaluation protocol and metrics.} We use the original public NOCS evaluation code for the pose evaluation, as is done in most prior work. A small bug in the NOCS code is later reported by \cite{lin2022sar}. We provide the updated results with the bug fixed in \supplementary. We use the Chamfer distance as the metric for the shape evaluation and follow the evaluation code from \cite{tian2020shape}. All the pose accuracy metrics in Tables \ref{tab: nocspose} and \ref{tab: ablation} are in terms of the percentage of the results that fall under the specified thresholds. IoU is the 3D overlap between the estimated 3D bounding box and the ground-truth 3D bounding box of the object. Most neural methods directly predict three scales along the three axes ($x,y,z$) for more accurate 3D bounding box prediction. To build the 3D bounding box, our method multiplies the estimated scale by the aspect ratios along the three axes of the estimated shape, whose diagonal is normalized to one (the final estimated scale absorbs the normalization factor).

\subsection{Results} \label{sec: results}
Table \ref{tab: nocspose} shows the benchmarking results on NOCS REAL. Our method achieves overall performance that surpasses or is comparable to many data-driven approaches. As an optimization-based approach, our method can reach very high accuracy once it finds the convergence basin around the ground-truth, which is evidenced by its dominating performance in the 5\degree 5cm and 5\degree 10cm metrics. The inability of our method to estimate three scales along the three axes is potentially the cause of the low IoU50 metric.

\input{includes/nocsbenchmark_short}
We emphasize the comparison between our method and ASM-Net \cite{akizuki2021asm} since we both adopt an ASM. Despite not using any pose-annotated data, our method surpasses ASM-Net, which leverages annotated data, by a significant margin. Our no-learning variation (w/o S.C.) is even stronger than their ICP-assisted version in the 5\degree 5cm metric and closely comparable in the IoU75 and 10\degree 5cm metrics. Moreover, we attain a dominant win against the previous no-learning method, Neural Sythesis-by-Analysis \cite{chen2020category}, and the previous mesh-based method, Metric Scale \cite{lee2021category}.

In Table \ref{tab: nocsshape}, our method provides moderate shape estimation accuracy, ranking in the middle among several learning-based methods. ASM-Net achieves better shape accuracy than ours. The point-based ASM used by ASM-Net is more flexible (due to the undefined connectivity between points) than our mesh-based ASM though it is less expressive in the geometry (due to the loss of surface). The main limitation for our mesh-based ASM comes from the template deformation step which restricts the mesh topology to the template topology, bringing in another source of error other than the latent shape reconstruction error. However, this limitation can be alleviated by manually engineering category-specific templates instead of a spherical template throughout. Additionally, our shape initialization network performs classification instead of regression. It has the advantage of being failure-safe since even if it makes a wrong prediction, the prediction is always a valid shape in the category. However, this merit comes at the expense of losing the ability to interpolate or extrapolate, resulting in lower accuracy.

\input{includes/nocsfigures_short}
In Fig. \ref{fig: nocsviz}, we visualize some example results. The estimated objects in general fit the depth measurements very well, showing the effectiveness of \algName~in minimizing the objective function \eqref{eq: shapeicp}. The errors mostly stem from the inherent ambiguity in the problem. For example, a can is easily fit upside down. In addition, our method notably tends to make mistakes on mugs and cameras. The orientation of these objects mostly depends on local features such as the mug handle. Therefore, the global minimum for the orientation has a very small convergence basin, which is extremely difficult to find under the influence of noise and partial views. In general, our algorithm has a tendency to fit the bulk of the object well but ignore the local details. \diffcolor{More explanations are in \supplementary.} 

The current focus of our method is on feasibility and accuracy, while a discussion of runtime is in \supplementary.

\subsection{Ablation}
\input{includes/ablation}
We ablate the modules for local minimum handling introduced in Section \ref{sec: copemin} in Table \ref{tab: ablation}. The symmetry check provides uniform improvements across the metrics. The EM treatment and the depth rendering check lead to more notable improvements in the higher accuracy metrics, such as 5\degree 5cm and 5\degree 2cm. The proposed multi-hypothesis estimation significantly outperforms single-hypothesis estimation that uses $\mathcal{S}_r + \mathcal{S}_\sigma$ to select the best hypothesis at the beginning of the iteration. More analysis is in \supplementary.

\subsection{\algName~versus Learning-based Methods}
As discussed in Section \ref{sec: results}, due to the inherent ambiguity in the problem, the estimated pose and shape may fit the depth measurements very well (i.e. achieving a low objective value or even the global minimum) but are different from the ground-truth. This is a limitation of the objective function. Another limitation of our \algName~solver is its inability to find a very small convergence basin. These limitations are where learning-based methods can help, since  they essentially solve a mapping problem from inputs to outputs instead of an optimization problem. However, our optimization-based approach shines when it finds the convergence basin, it can achieve very high accuracy. Furthermore, our method can be quickly deployed in new environments without curating an object pose dataset for training.

\section{Conclusion}
We proposed the \algName~algorithm for category-level object pose and shape estimation from a single depth image. Our method implements a novel mesh-based active shape model (ASM) to represent categorical object shape. Remarkably, the method does not require pose-annotated data, unlike the vast majority of the methods in the field. However, the performance of \algName~surpasses or is on par with many learning-based methods that rely on pose-annotated data. Future directions include how to better disambiguate the problem, potentially by exploring RGB-image-based constraints, and how to better initialize the algorithm closer to the convergence basin of a global minimum.

\bibliographystyle{IEEEtran}
\bibliography{shapeICP_zyh}

\newpage
\clearpage
\appendix
\section*{\diffcolor{Mesh-Based ASM Versus Point-Based ASM and Our Contribution}}
\diffcolor{We start with a detailed comparison between the point-based ASM and our mesh-based ASM. The three requirements to build a point-based ASM are stated in \cite{akizuki2021asm}:
\begin{enumerate}
  \item The database models have the same number of points. This is achieved by simply sampling the same number of points on the ShapeNet \cite{chang2015shapenet} models.
  \item The database models are in the same canonical pose. This is trivial because the ShapeNet models are pre-oriented to be in the same canonical pose.
  \item Semantically identical point correspondences exist across the database models. In other words, each dimension of the PCA input feature vectors should have consistent semantic meaning. For example, one dimension can be the top center point of a bottle cap. Then all the feature vectors should roughly have this dimension to mean this top center point. This semantic consistency in each dimension also implies the same feature vector length, i.e. the same number of points. The authors of \cite{akizuki2021asm} find the correspondences by using an off-the-shelf non-rigid point registration method to register a held-out model to all the other models.
\end{enumerate}
Our mesh-based ASM would be a point-based ASM if we ignored or removed the connectivity information, the edge set $\mathcal{E}$ and the face set $\mathcal{F}$, because only the vertices (points) would be left in this way. Therefore, to build our mesh-based ASM, one additional constraint is needed besides the three requirements for the point-based ASM (with points switched to vertices in the context of mesh-based ASM). That is the need to retain the same connectivity ($\mathcal{E}, \mathcal{F}$) across the database models. This additional constraint leads to a different methodology. In summary, for our mesh-based ASM,
\begin{enumerate}
  \item The database models have the same number of vertices. We use deformed templates to represent database models, so the number of vertices is always equal to the number of vertices in the template.
  \item The database models are in the same canonical pose. This is trivial because the ShapeNet models are pre-oriented to be in the same canonical pose.
  \item Semantically identical vertex correspondences exist across the database models. The projective-style deformation process, the models being pre-oriented to the same canonical pose, and the models being in the same category cause the same template vertex to be deformed to roughly the same semantic area across the models.
  \item The connectivity ($\mathcal{E}, \mathcal{F}$) is consistent across the database models, which is achieved because we use the deformed templates to represent the models and they all have the same connectivity as the template. 
\end{enumerate}
Our contribution is to provide this methodology to build a mesh-based object ASM. The deformation losses and process used in our paper are considered prior work \cite{gkioxari2019mesh, wang2018pixel2mesh, ravi2020pytorch3d}. However, we write out the losses for clarity since the same deformation losses are used later in the shape estimation step. In addition, the use of a mesh-based ASM (rather than a point-based ASM) is also novel in the task of object pose and shape estimation.}

\section*{Visualization of EM}
\begin{figure}[h]
\centering
\begin{subfigure}{0.49\linewidth}
    \includegraphics[width=1.0\columnwidth]{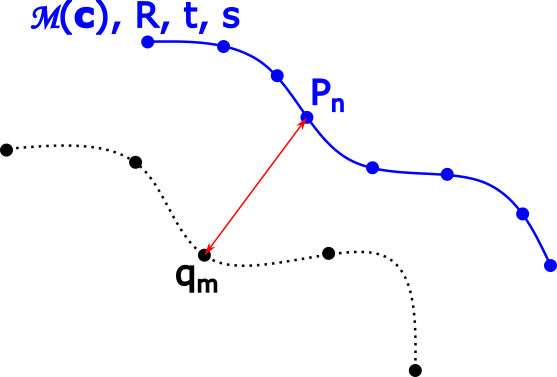}
\end{subfigure}
\begin{subfigure}{0.49\linewidth}
    \includegraphics[width=1.0\columnwidth]{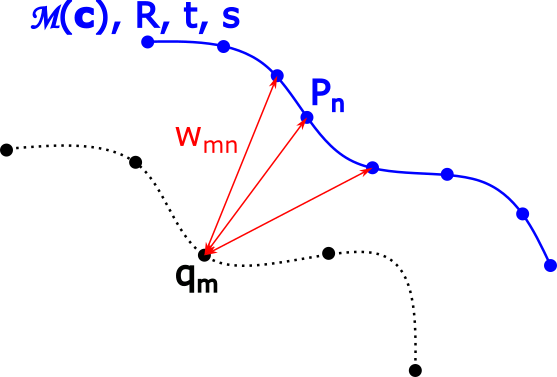}
\end{subfigure}
\caption{Visualization of the original hard one-to-one association (left) and the EM soft associations for $Q=3$ (right). $\mathcal{M}(\shapec)$ is the ASM model that returns a mesh. The blue curve represents the mesh surface transformed by the rotation $R$, translation $t$, and scale $s$. The black dotted curve is the surface of the measured object.}
\label{fig: emviz}
\end{figure}

Fig. \ref{fig: emviz} visualizes the soft associations in \eqref{eq: emshapeicp}.

\section*{Derivation of EM}
We begin by stating our modeling choices and assumptions. First, each measurement $m$ is accompanied by a latent variable $z_m$ that models the association of this measurement. Specifically, 
\begin{equation}
    \mathcal{P}(z_m = n) = 
    \begin{cases}
        \frac{1}{Q} & \text{if $n\in\mathcal{N}_Q(q_m)$} \\
        0           & \text{otherwise}
    \end{cases}
\label{eq: emass}
\end{equation}
where $\mathcal{P}(z_m = n)$ is the probability of $q_m$ associated to $p_n$ and $\mathcal{N}_Q(q_m)$ finds the $Q$ closest transformed model points to $q_m$ (i.e. the top $Q$ solutions to $\arg\min_{n}{\|sRp_n + t - q_m\|_2^2}$). We further model the conditional data likelihood as
\begin{equation}
    \mathcal{P}(q_m|z_m = n) = \mathsf{N}(sRp_n + t - q_m; 0, \Sigma_m)
\label{eq: xgivenz}
\end{equation}
where $\mathsf{N}(\cdot; 0, \Sigma_m)$ is a zero-mean Gaussian distribution with covariance $\Sigma_m$ (i.e. conditioned on $q_m$ associated to $p_n$, the residual follows a Gaussian distribution). We assume independence $q_m|z_m \perp z_{m'}$, $q_m|z_m \perp q_{m'}|z_{m'}$ and $z_m \perp z_{m'}$, and isotropic covariance $\Sigma_m = \mathsf{diag}(\sigma_m^2)$.

We now explain these modeling choices \eqref{eq: emass}\eqref{eq: xgivenz} and assumptions. Since our sensor is a depth camera, the correct association for a depth measurement at a pixel should be made by ray casting instead of looking for the closest point to $q_m$. Therefore, $\mathcal{N}_Q(q_m)$ in \eqref{eq: emass} should ideally find the $Q$ closest points near the point of the first hit of the cast ray ($Q$ points instead of one to account for the association uncertainties caused by the errors in the object pose and shape). However, the approximation of the Euclidean closest points near $q_m$ greatly simplifies the computation and has shown robustness in ICP. Likewise, \eqref{eq: xgivenz} is also an approximation to simplify the computation. When the association is given for a pixel in a depth image, the uncertainty should only be along the ray through the camera optical center, the pixel, and the associated point (1D uncertainty). By using the 3D Gaussian approximation as in \eqref{eq: xgivenz}, we assume the sensor is not a depth camera but rather the following generative process that directly makes the spatial measurement:
\begin{equation}
    q_m = sRp_{z_m} + t + \epsilon_m
\label{eq: qgenproc}
\end{equation}
where $\epsilon$ is Gaussian noise following $\mathsf{N}(0, \Sigma_m)$.

The independence assumption $q_m|z_m \perp z_{m'}$ (i.e. $\mathcal{P}(q_m|z_m, z_{m'}) = \mathcal{P}(q_m|z_m)$) is reasonable. If the measurement association is already given, the association of another measurement should provide no extra information. Similarly, $q_m|z_m \perp q_{m'}|z_{m'}$ is also reasonable. Given the associations, two measurements are assumed to be independent (unless there is some systematic bias of the sensor). The assumption of all the latent variables ($z_m$'s) being independent of each other is not totally valid for a depth camera because the association at one pixel gives away the ray orientation (i.e. the bearing of the camera along that ray) and the association of the next pixel can only be made on a cone (possibly an oblique circular cone if the first pixel is not on the principal axis) with its center line being the ray given by the first association (the ray of the next pixel starts from the same optical center but rotates around the first ray as the camera rotates while the next pixel keeps the same distance to the first pixel on the image place). Nevertheless, for the generative model \eqref{eq: qgenproc}, it is reasonable to assume that the association of one measurement does not provide information to another association ($z_m$'s are independent of each other. i.e. $z_m \perp z_{m'}$) and it also greatly simplifies the derivation of EM.

To derive the EM formulation, we begin with the data likelihood: 
\begin{equation}
\begin{split}
    \mathcal{P}(q_m) & = \sum_n{\mathcal{P}(q_m|z_m = n)\mathcal{P}(z_m=n)}\\
    & = \sum_{n\in\mathcal{N}_Q(q_m)}{\frac{1}{Q}\mathsf{N}(sRp_n + t - q_m; 0, \Sigma_m)}
\end{split}
\label{eq: px}
\end{equation}
Making use of the independence assumptions and denoting $\mathcal{Z} = \{z_m\}_{m=1}^M$,
\begin{equation}
\begin{split}
    \mathcal{P}(\{q_m\}_{m=1}^M|\mathcal{Z}) & = \prod_{m=1}^M{\mathcal{P}(q_m|\mathcal{Z})}\\
    & = \prod_{m=1}^M{\mathcal{P}(q_m|z_m)}\\
    & = \prod_{m=1}^M{\mathsf{N}(sRp_{z_m} + t - q_m; 0, \Sigma_m)}
\end{split}
\end{equation}
where the first equality assumes $\mathcal{P}(q_m|z_m) \perp \mathcal{P}(q_{m'}|z_{m'})$ and the second equality assumes $\mathcal{P}(q_m|z_m, z_{m'}) = \mathcal{P}(q_m|z_m)$. Since we have further assumed all the latent variables ($z_m$'s) are independent of each other, it leads to
\begin{equation}
    \mathcal{P}(\mathcal{Z}) = 
    \begin{cases}
        \frac{1}{Q^M} & \text{if $z_m\in\mathcal{N}_Q(q_m)~\forall m$} \\
        0           & \text{otherwise}
    \end{cases}
\end{equation}
We can now compute the posterior probability of the latent variables:
\begin{equation}
\begin{split}
    \mathcal{P}(\mathcal{Z}|\{q_m\}_{m=1}^M) & = \frac{\mathcal{P}(\mathcal{Z}, \{q_m\}_{m=1}^M)}{\mathcal{P}(\{q_m\}_{m=1}^M)}\\
    & = \frac{\mathcal{P}(\{q_m\}_{m=1}^M|\mathcal{Z})\mathcal{P}(\mathcal{Z})}{\sum_Z{\mathcal{P}(\{q_m\}_{m=1}^M|\mathcal{Z})\mathcal{P}(\mathcal{Z})}}\\
\end{split}
\end{equation}
whose result is
\begin{equation}
\begin{cases}
    \frac{\frac{1}{Q^M}\prod_{m=1}^M{\mathsf{N}(sRp_{z_m} + t - q_m; 0, \Sigma_m)}}{\sum_\mathcal{Z}{\frac{1}{Q^M}\prod_{m=1}^M{\mathsf{N}(sRp_{z_m} + t - q_m; 0, \Sigma_m)}}} & \text{if $z_m\in\mathcal{N}_Q(q_m)~\forall m$} \\
    0 & \text{otherwise}
\end{cases}
\end{equation}
In the expectation step, we need to compute $\sum_\mathcal{Z}{\hat{\mathcal{P}}(\mathcal{Z}|\{q_m\}_{m=1}^M)\ln{\mathcal{P}(\mathcal{Z}, \{q_m\}_{m=1}^M)}}$, where $\hat{\mathcal{P}}(\mathcal{Z}|\{q_m\}_{m=1}^M)$ is given a hat notation to indicate that it is computed from the last available estimates \cite{bishop2006pattern}:
\begin{multline}
    \sum_\mathcal{Z}{\hat{\mathcal{P}}(\mathcal{Z}|\{q_m\}_{m=1}^M)\ln{\mathcal{P}(\mathcal{Z}, \{q_m\}_{m=1}^M)}} \\
    = \frac{1}{\sum_\mathcal{Z}{\frac{1}{Q^M}\prod_{m=1}^M{\mathsf{N}(\hat{s}\hat{R}\hat{p}_{z_m} + \hat{t} - q_m; 0, \Sigma_m)}}} \\
    \sum_\mathcal{Z}\Biggl\{\frac{1}{Q^M}\biggl[\prod_{m=1}^M{\mathsf{N}(\hat{s}\hat{R}\hat{p}_{z_m} + \hat{t} - q_m; 0, \Sigma_m)}\biggl] \\
    \sum_{m=1}^M{\ln{\frac{1}{Q}} + \ln{\mathsf{N}(sRp_{z_m} + t - q_m; 0, \Sigma_m)}}\Biggl\}
\label{eq: estep}
\end{multline}
where $\sum_\mathcal{Z}$ is over all possible combinations of $\{z_m\in\mathcal{N}_Q(q_m)\}_{m=1}^M$. To simplify \eqref{eq: estep}, we examine the pattern in a simple case. Suppose we have three measurements ($M=3$) and we use $a$, $b$, $c$ to denote the three $\frac{1}{Q}\mathsf{N}(\hat{s}\hat{R}\hat{p}_{z_m} + \hat{t} - q_m; 0, \Sigma_m)$ terms. We then use $a'$, $b'$, $c'$ to denote the three $\ln{\frac{1}{Q}} + \ln{\mathsf{N}(sRp_{z_m} + t - q_m; 0, \Sigma_m)}$ terms. Suppose for each measurement, we have two possible associations (i.e. 2 realizations of $z_m$ and 8 realizations of $\mathcal{Z}$ for 3 measurements) which are denoted by subscript 1 and 2. The normalization term $\sum_\mathcal{Z}{\frac{1}{Q^M}\prod_{m=1}^M{\mathsf{N}(\hat{s}\hat{R}\hat{p}_{z_m} + \hat{t} - q_m; 0, \Sigma_m)}}$ becomes
\begin{gather*}
    a_1b_1c_1 + a_2b_1c_1 + a_1b_2c_1 + a_2b_2c_1\\
    + a_1b_1c_2 + a_2b_1c_2 + a_1b_2c_2 + a_2b_2c_2
\end{gather*}
The remaining term becomes
\begin{gather*}
    a_1b_1c_1(a_1' + b_1' + c_1') + a_2b_1c_1(a_2' + b_1' + c_1')\\
    + a_1b_2c_1(a_1' + b_2' + c_1') + a_2b_2c_1(a_2' + b_2' + c_1')\\
    + a_1b_1c_2(a_1' + b_1' + c_2') + a_2b_1c_2(a_2' + b_1' + c_2')\\
    + a_1b_2c_2(a_1' + b_2' + c_2') + a_2b_2c_2(a_2' + b_2' + c_2')
\end{gather*}
If we look at the coefficient of $a_1'$, it is $a_1(b_1c_1 + b_2c_1 + b_1c_2 + b_2c_2)$ which is canceled out with the normalization term to only leave $\frac{a_1}{a_1+a_2}$. Similarly, the coefficient of $a_2'$ is $\frac{a_2}{a_1+a_2}$, the coefficient of $b_2'$ is $\frac{b_2}{b_1+b_2}$, and so forth. If we gather the terms, \eqref{eq: estep} in this simple case is simplified to
\begin{gather*}
    \frac{a_1a_1'}{a_1+a_2} + \frac{a_2a_2'}{a_1+a_2} + \frac{b_1b_1'}{b_1+b_2}\\
    + \frac{b_2b_2'}{b_1+b_2} + \frac{c_1c_1'}{c_1+c_2} + \frac{c_2c_2'}{c_1+c_2}\\
    = \frac{a_1a_1'+a_2a_2'}{a_1+a_2} + \frac{b_1b_1'+b_2b_2'}{b_1+b_2} + \frac{c_1c_1'+c_2c_2'}{c_1+c_2}
\end{gather*}
We recognize that there is an outer summation over the measurements (one term each for $a$, $b$, and $c$), and for each measurement term in the summation, both the denominator and the numerator are summations over the associations (realizations of $z_m$ indicated by subscripts 1 and 2). With this pattern, \eqref{eq: estep} in general is simplified to
\begin{multline}
    \sum_{m=1}^M\Biggl\{\frac{1}{\sum_{z_m\in\mathcal{N}_Q(q_m)}{\frac{1}{Q}\mathsf{N}(\hat{s}\hat{R}\hat{p}_{z_m} + \hat{t} - q_m; 0, \Sigma_m)}}\\
    \sum_{z_m\in\mathcal{N}_Q(q_m)}\biggl[\frac{1}{Q}\mathsf{N}(\hat{s}\hat{R}\hat{p}_{z_m} + \hat{t} - q_m; 0, \Sigma_m)\\
    \Bigl(\ln{\frac{1}{Q}} + \ln{\mathsf{N}(sRp_{z_m} + t - q_m; 0, \Sigma_m)}\Bigl)\biggl]\Biggl\}
\label{eq: simpleestep}
\end{multline}
Assuming isotropic covariance $\Sigma_m = \mathsf{diag}(\sigma^2_m)$ and considering only the terms that involve $sRp_{z_m} + t$ (other terms are irrelevant in the maximization), we obtain \eqref{eq: emshapeicp} by maximizing \eqref{eq: simpleestep}.

\section*{Implementation Details}
We present all the parameter settings in our algorithm in Table \ref{tab: params}. \diffcolor{We briefly explain our choice of the number of bases ($K$) in our ASM. There are a few factors that affect this number $K$, including the number of models available for the category, the intra-category variations, and how detailed the ASM has to be for high reconstruction accuracy. In practice, we look at the magnitudes of the singular values and the percentage of variance explained, which are returned by PCA. They should provide a good indicator for the first two factors. We adopt the common technique to look for the elbow point where the curve (e.g. singular value magnitude versus $K$) starts to flatten out. In addition, we compute the reconstruction errors using different $K$'s on ShapeNetCore \cite{chang2015shapenet} and pick the $K$ with the reconstruction error well below the baseline shape estimation error on the NOCS REAL \cite{wang2019normalized} benchmark so that our shape estimation error is not dominated by the ASM reconstruction error. After taking into account all the metrics above, we try to choose the smallest $K$ possible to be computationally efficient. In a later section, we perform a sensitivity analysis of the parameters for the multi-hypothesis estimation and shed light on our choice of the parameter values.}
\input{includes/params}

\section*{Per-Class Shape Evaluation Results}
The per-class results are in Table \ref{tab: appnocsshape}.
\input{includes/nocsbenchmark_shape}

\section*{Runtime}
The current algorithm without code optimization runs on an Nvidia TITAN RTX GPU at about 8 seconds per object instance, which can be inferior to most learning-based methods at inference time. The runtime may be acceptable in an online setting if the algorithm is run only on a sparse set of key-frames for static objects. Table \ref{tab: runtime} shows an ablation analysis for runtime. We observe that reducing the system to run on a single hypothesis (or initial guess) does not significantly reduce the runtime, while skipping the shape estimation step greatly reduces the runtime. A runtime breakdown for the most time-consuming processes in the shape estimation step is given in Table \ref{tab: shaperuntime}. There is no dominating sub-component (except the back-propagation step). In general, the loss computation (except for the KNN-finding, i.e. computing $\mathcal{N}_Q(q_m)$, and the Chamfer loss \eqref{eq: emshapeicp}, which together take only 0.7\%) and the back-propagation (i.e. loss gradient) take the largest share. Therefore, the shape step after the correspondence finding (i.e. KNN-finding) may be potentially sped up by formulating it as a linear least squares problem (since the ASM is linear) with proper regularization losses and computing the closed-form solution.
\input{includes/runtime}

\section*{Ablation}
We perform more detailed ablation analysis to study the effects of different shape initialization approaches and the performance limit of our pose and shape estimation method. In all these experiments, we used the ground-truth object mask (instead of Mask RCNN) for consistency when controlling different variables. In the shape initialization experiments, we initialized the shape code with $\shapec = 0$ (mean shape across a category), the mean shape code across a category $\shapec = \frac{1}{U}\sum_u{\tilde{\shapec}_u}$, the network-classified closest shape $\shapec = \tilde{\shapec}_{u^*}$, and the ShapeNet \cite{chang2015shapenet} ground-truth closest shape $\shapec = \shapec_{u^*}$, which is the shape in ShapeNet closest to the ground-truth shape by the Chamfer distance. The results of running these different initialization approaches are given in Table \ref{tab: shapeinitabl}.
\input{includes/shapeinitabl}

The overall trend of the results in Table \ref{tab: shapeinitabl} meets our expectation. Initializing with the mean shape or mean code has nearly identical results. Initializing with the neural network classified closest shape has much better performance than initializing with the mean shape (or mean code). This shows the advantage of leveraging the power of prior data captured by the network. We further imagine an optimal classifier which returns the ground-truth closest ShapeNet shape for initialization. The performance of this optimal classifier is better than the network classifier by a small margin in pose estimation but a larger margin in shape estimation. This suggests that the optimal classifier indeed provides better shape initialization. However, the pose estimation is somewhat insensitive to the improvement in shape.

We are also interested in how the shape accuracy affects the pose estimation performance. In this set of experiments, we fix the shape to various sources without optimizing the shape, so only the pose estimation is run given different levels of shape accuracy. The tested shape sources are the categorical mean shape, ground-truth shape, closest ShapeNet shape, closest deformed template, closest PCA shape. To find the ground-truth closest shape, we match every ground-truth shape to a ShapeNet model using the Chamfer distance. After the closest shape is found, we use the original ShapeNet model, its deformed template (Section \ref{sec: asm}), and the PCA reconstruction of the deformed template (with $K$ principal components) as the given shape to run the pose estimation. The results are summarized in Table \ref{tab: shapeeffpose}.
\input{includes/shapeeffectsonpose}

Since we do not optimize the shape, the Chamfer distance in Table \ref{tab: shapeeffpose} directly reflects the shape accuracy of the various shape sources. The overall trend shows that a more accurate shape leads to better pose accuracy. However, when the given shapes have similar accuracy, the less accurate shape (e.g. the PCA reconstructed shape) results in higher pose estimation performance. In particular, the ground-truth shape does not achieve the best performance. This agrees with our discussions in Section \ref{sec: results}. Our algorithm does not rely on fine geometric details to fit the model to the measured depth points, since the depth measurements are both very noisy and partial, and consequently, they are also not reliable in the details. Given the smoother PCA reconstruction, the bulk of the object is fit, which smoothens the optimization landscape. The ultra-fine ground-truth shape may only create more pitfalls (local minima) in the optimization landscape.

Comparing between Table \ref{tab: shapeinitabl} and Table \ref{tab: shapeeffpose} (Mean Shape vs Mean Shape, CNN Closest vs CNN Closest PCA, GT Closest vs GT Closest PCA), we observe that the shape optimization step does improve both the shape and the pose when the initial shape is relatively poor (Mean Shape vs Mean Shape, CNN Closest vs CNN Closest PCA). When the initial shape is already accurate (GT Closest vs GT Closest PCA), including the shape step is not very effective, probably because the noisy and partial depth measurements do not bring in additional useful information.

Lastly, we examine the shape estimation performance (initialized by the neural network classification) given and not given the ground-truth pose. When the ground-truth pose is given and not optimized, the shape estimation has the mean (across the categories) Chamfer distance ($10^{-3}$) 3.18. When the pose is estimated, the mean Chamfer distance is 2.39. We again hypothesize that the noisy and partial depth measurements actually cause the shape to fit to noise if the pose is fixed at the ground-truth. It is better to absorb some of the errors into the pose estimation for better shape estimation accuracy.

\section*{\diffcolor{Sensitivity Test for Multi-Hypothesis Estimation}}
\label{sec: sensitivity_multihyp}
\diffcolor{It is important to understand how the parameters in the multi-hypothesis estimation could affect the final results. We perform sensitivity tests on two parameters, the number of hypotheses and the iteration numbers where down-selection happens (Table \ref{tab: sens_multihyp}). A relatively clear trend is that the accuracy improves and the runtime increases when we increase the number of hypotheses, which is expected. However, we can see that decreasing the number of hypotheses by half has a more significant detrimental impact than the benefits brought by doubling the number of hypotheses. In other words, our default setting seems optimal since increasing the number of hypotheses brings marginal improvements at the cost of large runtime increase, while decreasing it hurts the performance too much. The results of changing the down-selection iteration numbers show that our default setting seems again optimal with either direction being more or less worse in certain metrics. Overall, we try to balance the runtime and accuracy in the default setting.}
\input{includes/sensitivity_multihyp}

\section*{\diffcolor{Explanations of the Limitations}}
\diffcolor{We mentioned two major limitations of our approach: dealing with the natural ambiguity in the problem and dealing with a small convergence basin. We provide more detailed explanations here of these two limitations.}
\subsection{\diffcolor{Ambiguity}}
\diffcolor{A good example is Fig. 3, where the rightmost image is the result of fitting the model to the measured point cloud which is back-projected from the segmented areas in the depth image. The center image is the segmentation results from Mask RCNN \cite{vuola2019mask}. If we back-project the depth points in the segmented areas for both the can and the laptop, we obtain the blue point cloud in the rightmost image. The point cloud for the laptop is extremely fragmented, which cause a lot of ambiguity when fitting the model to the point cloud. The ambiguity eventually manifests itself in a way that the fitting seems okay but the fitted laptop is not where it should be, and in this case, it is in collision with the can. The purpose of Fig. 3 is to provide readers with an example of the ambiguity that often exist in this type of problem. Another simpler example is when we have a front view of a laptop and its thickness is not observed, a thicker laptop model and a thinner laptop model could fit equally well to the measurements but only one of them matches the ground-truth.}
\subsection{\diffcolor{Small Convergence Basin}}
\begin{figure}[ht]
\begin{center}
    \begin{subfigure}{0.3\textwidth}
        \includegraphics[width=1.0\columnwidth]{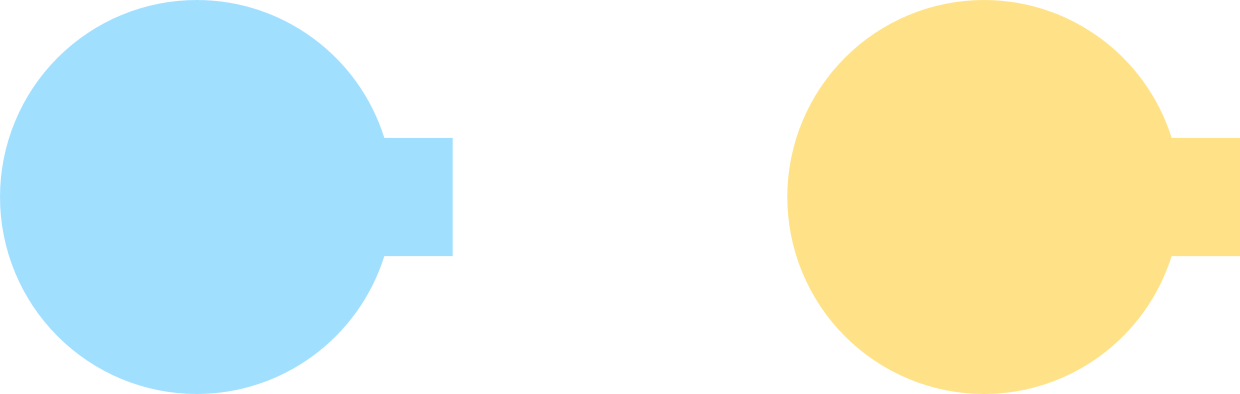}
    \caption{Source model (blue) and target model (yellow).}
    \end{subfigure}
\end{center}
\hfill
\begin{center}
    \begin{subfigure}{0.37\textwidth}
        \begin{overpic}[width=1.0\columnwidth, tics=5]{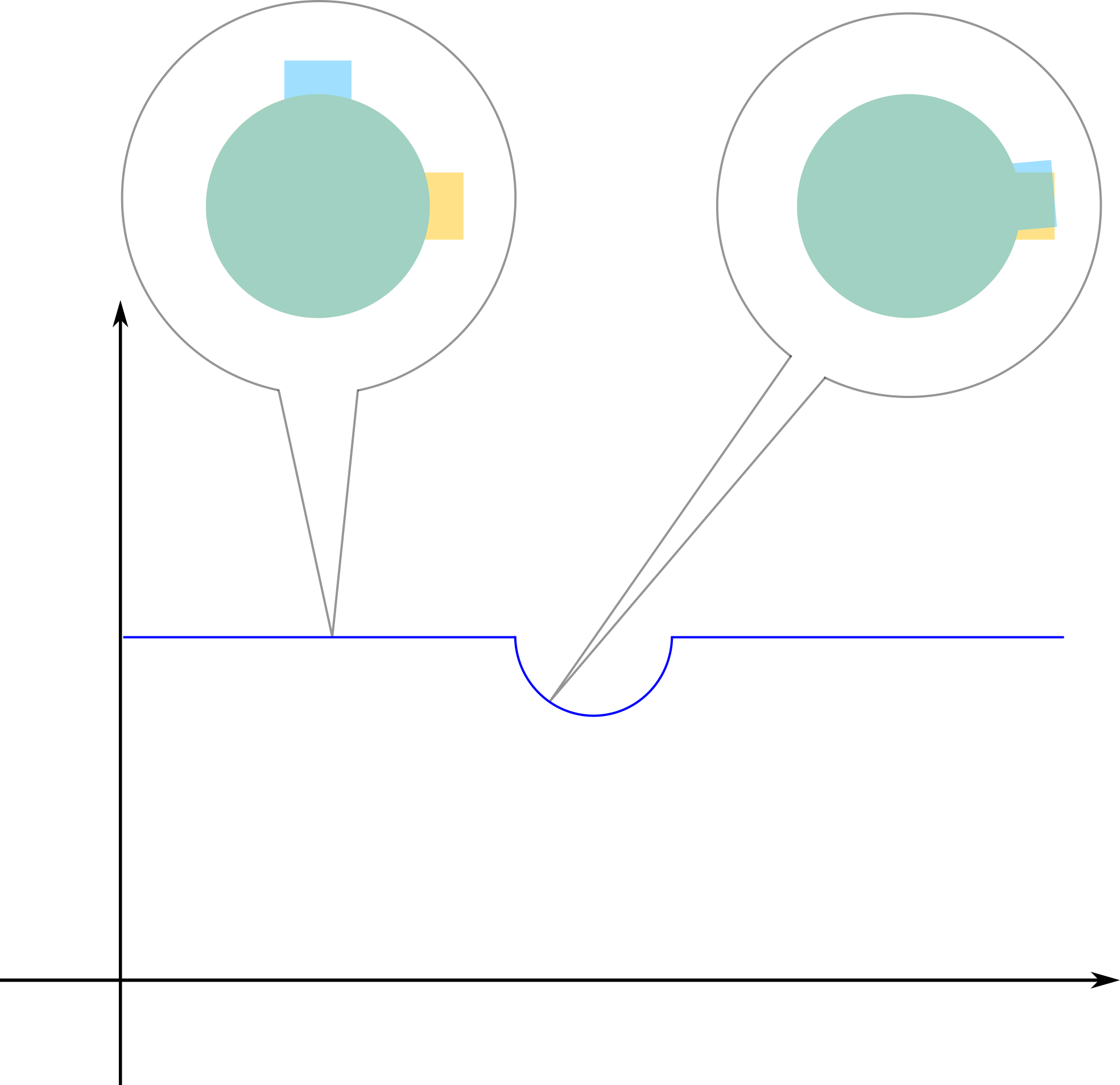}
            \put(-1,67){Cost}
            \put(92,4){Rotation}
        \end{overpic}
        \caption{Optimization landscape when fitting the source to the target assuming the only optimization variable is the in-plane rotation.}
        \label{fig: localfeature_obj}
    \end{subfigure}
\end{center}
\caption{\diffcolor{A schematic of the local feature issue and the optimization landscape.}}
\label{fig: localfeature}
\end{figure}

\diffcolor{Consider the following simplified problem in Figure \ref{fig: localfeature}, which can be seen as a top-down view of a mug. Suppose that the only optimization variable is the 1D in-plane rotation. The optimization landscape of our objective function (which is the same as the ICP objective here since the shape is fixed) would schematically look like Figure \ref{fig: localfeature_obj}. When the estimated handle is rotated far away from the handle in the target, the objective function is roughly flat, having zero gradient. Until the estimated handle is rotated near the target handle, there appears a ditch in the objective function. Such optimization landscape is hard to minimize because there is no gradient in the flat region and the convergence basin is very small in the solution space. Although this is a simplified case, it implies that the similar problem could exist for an object whose pose depends on local features rather than the bulk of the object shape. Examples are the camera class and the mug class. In our experiments, the problem is even aggravated due to measurement noise and partial views, which the simplified case here does not have.}

\section*{Updates on the Evaluation}
SAR-Net\footnote{\url{https://github.com/hetolin/SAR-Net}} pointed out a bug in the NOCS \cite{wang2019normalized} evaluation code\footnote{\url{https://github.com/hughw19/NOCS_CVPR2019}}. We found that this bug could impact the 3D IoU values. It had negligible effects on the other pose accuracy metrics. Most of the prior work used the original NOCS code for evaluation and so did we. However, we include the results obtained from the debugged code in Table \ref{tab: debugupdate} and \ref{tab: debugablation} for readers' reference.
\input{includes/debugged}

\end{document}

%% file: includes/fullalg.tex
\setlength{\textfloatsep}{1.5pt}
\setlength{\intextsep}{2pt}
\begin{algorithm}[h]
    \footnotesize
    \caption{\algName \label{alg: shapeicp}}
    \KwIn{back-projected depth points of the object $\{q_m\}_{m=1}^M$; background-masked depth image $D$; initial guesses $\hat{t}$, $\hat{s}$, and $\hat{\shapec}$; a set of G rotation guesses $\{\hat{R}_g\}_{g=1}^G$; an ASM with bases $\{b_k\}_{k=1}^K$; a set of symmetry operations $\{T_\psi\}_{\psi=1}^\Psi$}
    \KwOut{estimates of object pose and shape code $\hat{R}$, $\hat{t}$, $\hat{s}$, $\hat{\shapec}$}
    
    $\{\hat{t}_g, \hat{s}_g, \hat{\shapec}_g\} \leftarrow \{\hat{t}, \hat{s}, \hat{\shapec}\}~\forall~g$\;
    \For{$\text{\normalfont iter} = 1,2, \ldots, \text{\normalfont maxIterations}$  \label{line: iter}}{
        \If{\text{\normalfont down-select hypotheses}}{
            \For{$g = 1, \ldots, G$  \tcc{\footnotesize execute in parallel}}{
                $\mathcal{S}_{tot,g} = \mathcal{S}_{tot}(\hat{R}_g, \hat{t}_g, \hat{s}_g, \hat{\shapec}_g; \{q_m\}_{m=1}^M, D, \{T_\psi\}_{\psi=1}^\Psi)$\;
            }
            $\text{sort}(\{\hat{R}_g, \hat{t}_g, \hat{s}_g, \hat{\shapec}_g\}_{g=1}^G)~\text{by}~\{\mathcal{S}_{tot,g}\}_{g=1}^G$\;
            $\text{pick top hypotheses}~\{\hat{R}_g, \hat{t}_g, \hat{s}_g, \hat{\shapec}_g\}_{g=1}^{G_{new}}$\;
            $G \leftarrow G_{new}$\;
        }
        \For{$g = 1, \ldots, G$ \tcc{\footnotesize execute in parallel}} {
            $\text{find the }Q\text{ closest points }\mathcal{N}_Q(q_m) ~\forall~q_m$;
            $\{\hat{R}_g, \hat{t}_g, \hat{s}_g\} \leftarrow \text{solve \eqref{eq: emshapeicp} by Umeyama}$\;
            \If{\text{\normalfont stop shape estimation}}{
                continue\;
            }
            \For{$\text{\normalfont step} = 1,2, \ldots, \text{\normalfont maxSteps}$}{
                $\text{find the }Q\text{ closest points }\mathcal{N}_Q(q_m) ~\forall~q_m$\;
                $\hat{\shapec}_g \leftarrow \text{minimize } \mathcal{L}_{sc}~\text{by one step of SGD}$\;
                \tcc{\footnotesize \eqref{eq: emshapeicp} instead of \eqref{eq: shapeicp} in $\mathcal{L}_{sc}$}
            }
        }
    }
    \Return{$\{\hat{R}_{g=1}, \hat{t}_{g=1}, \hat{s}_{g=1}, \hat{\shapec}_{g=1}\}$}
\end{algorithm}
\vspace{-0.3em}

%% file: includes/nocsbenchmark_short.tex
\begin{table*}
\begin{center}
\begin{threeparttable}
\vspace{0.6em} 
\caption{Benchmarking results on NOCS REAL \cite{wang2019normalized}. Baseline results are from sources \cite{irshad2022shapo, akizuki2021asm}.  P.D. stands for pose-annotated data required for training and S.D. stands for shape-annotated image data \diffcolor{(but not including a database of categorical shape models). Inputs are the main inputs for inference.} S.C. means shape classification. The results are roughly in the order of overall performance. The best result in each metric is in bold. \diffcolor{\textsuperscript{1} pose annotated images are required but synthetically generated \textsuperscript{2} depth value for one pixel per object}}
\label{tab: nocspose}
{\footnotesize
    \begin{tabular}{|l| c c c | c c c c c c c|}
        \hline
        Method & P.D. & S.D. & \diffcolor{Inputs} & IoU25 & IoU50 & IoU75 & 5\degree 5cm & 5\degree 10cm & 10\degree 5cm & 10\degree 10cm \\
        \hline
        \hline
        NOCS \cite{wang2019normalized} & \checkmark & \checkmark & \diffcolor{RGB-D} & \bf{84.8} & 78.0 & 30.1 & 10.0 & 9.8 & 25.2 & 25.8 \\
        \hline
        Synthesis \cite{chen2020category} & \diffcolor{\checkmark\tnote{1}} & \hfill & \diffcolor{RGB} & -- & -- & -- & 0.9 & 1.4 & 2.4 & 5.5 \\
        \hline
        Metric Scale \cite{lee2021category} & \checkmark & \checkmark & \diffcolor{RGB-OD\tnote{2}} & 81.6 & 68.1 & -- & 5.3 & 5.5 & 24.7 & 26.5 \\
        \hline
        ASM-Net w/o ICP \cite{akizuki2021asm} & \checkmark\diffcolor{\tnote{1}} & \hfill & \diffcolor{Depth} & -- & 64.4 & 31.7 & 12.4 & -- & 37.7 & -- \\
        \hline
        \bf{\algName (Ours) w/o S.C.} & \hfill & \hfill & \diffcolor{Depth} & 82.6 & 58.3 & 37.0 & 32.4 & 33.2 & 42.2 & 43.4 \\
        \hline
        ASM-Net w/ ICP \cite{akizuki2021asm} & \checkmark\diffcolor{\tnote{1}} & \hfill & \diffcolor{Depth} & -- & 68.3 & 37.4 & 25.6 & -- & 43.7 & -- \\
        \hline
        \bf{\algName (Ours) w/ S.C.} & \hfill & \checkmark & \diffcolor{Depth} & 82.1 & 58.0 & 42.2 & \bf{36.5} & \bf{37.1} & 50.6 & 51.4 \\
        \hline
        ShapePrior \cite{tian2020shape} & \checkmark & \checkmark & \diffcolor{RGB-D} & 81.2 & 77.3 & \bf{53.2} & 21.4 & 21.4 & 54.1 & 54.1 \\
        \hline
        CenterSnap \cite{irshad2022centersnap} & \checkmark & \checkmark & \diffcolor{RGB-D} & 83.5 & \bf{80.2} & -- & 27.2 & 29.2 & \bf{58.8} & \bf{64.4} \\
        \hline
    \end{tabular}
}
\end{threeparttable}
\end{center}
\vspace{-1.0em}
\end{table*}

\begin{table}
\footnotesize
\begin{center}
\caption{Shape estimation results on NOCS REAL \cite{wang2019normalized} evaluated with the Chamfer distance ($10^{-3}$). The baseline results are from sources \cite{irshad2022shapo, akizuki2021asm}. See \supplementary~for the per-class results.}
\label{tab: nocsshape}
{\footnotesize
    \begin{tabular}{|l| c c | c |}
        \hline
        Method & P.D. & S.D. & Mean \\
        \hline
        \hline
        Categorical Prior \cite{tian2020shape} & \hfill & \hfill & 4.4 \\
        \hline
        \bf{\algName (Ours) w/o S.C.} & \hfill & \hfill & 3.8 \\
        \hline
        ShapePrior \cite{tian2020shape} & \checkmark & \checkmark & 3.2 \\
        \hline
        \bf{\algName (Ours) w/ S.C.} & \hfill & \checkmark & 2.6 \\      
        \hline
        CenterSnap \cite{irshad2022centersnap} & \checkmark & \checkmark & 1.5 \\
        \hline
        ASM-Net \cite{akizuki2021asm} & \checkmark & \hfill & 0.29 \\
        \hline
    \end{tabular}
}
\end{center}
\vspace{-1.5em}
\end{table}

%% file: includes/nocsfigures_short.tex
\begin{figure*}[ht]
    \centering
    \begin{subfigure}{0.18\linewidth}
        \includegraphics[width=1.0\columnwidth]{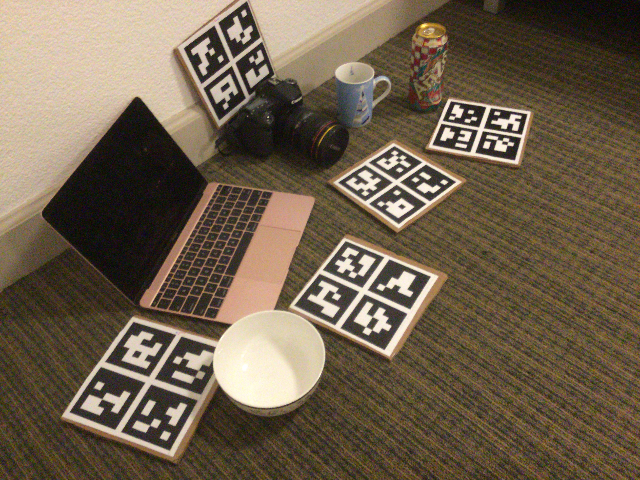}
    \end{subfigure}
    \begin{subfigure}{0.18\linewidth}
        \includegraphics[width=1.0\columnwidth]{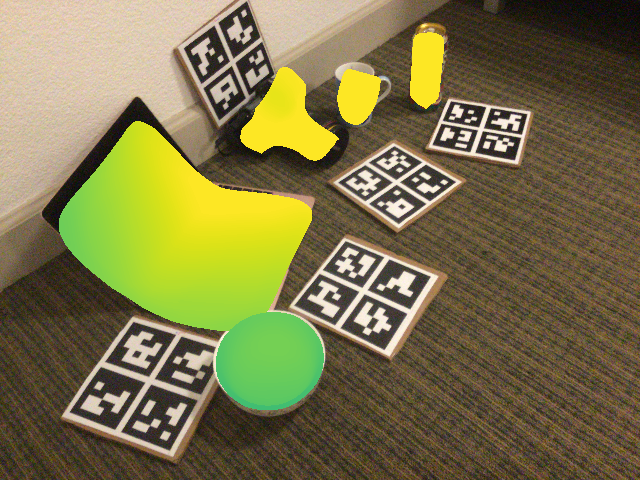}
    \end{subfigure}
    \begin{subfigure}{0.18\linewidth}
        \includegraphics[width=1.0\columnwidth]{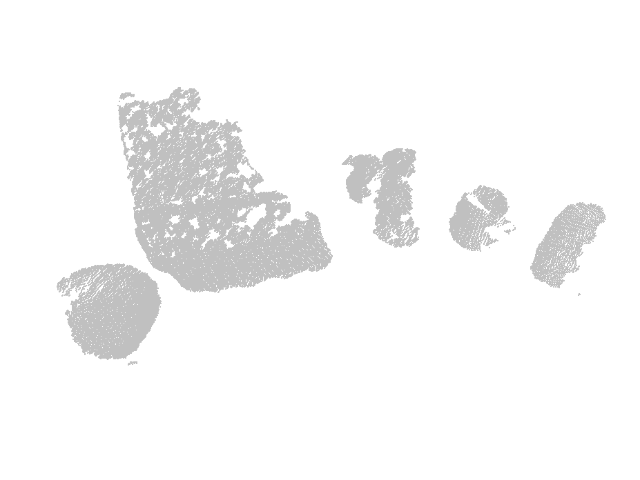}
    \end{subfigure}
    \begin{subfigure}{0.18\linewidth}
        \includegraphics[width=1.0\columnwidth]{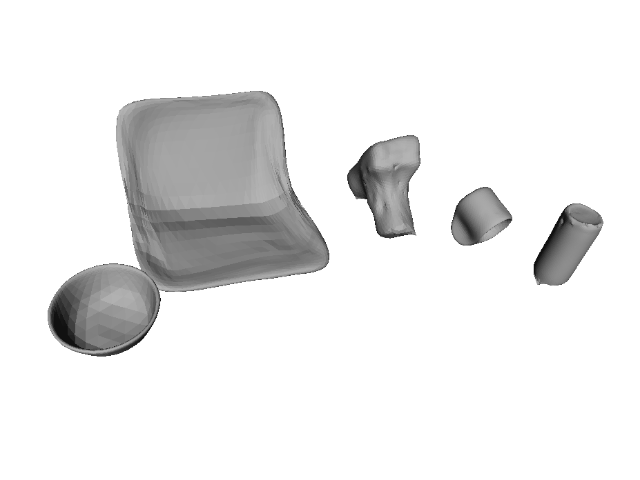}
    \end{subfigure}
    \par\smallskip
    \begin{subfigure}{0.18\linewidth}
        \includegraphics[width=1.0\columnwidth]{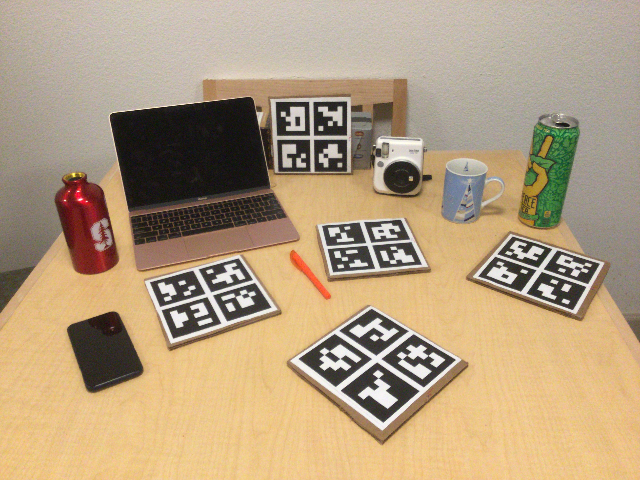}
    \end{subfigure}
    \begin{subfigure}{0.18\linewidth}
        \includegraphics[width=1.0\columnwidth]{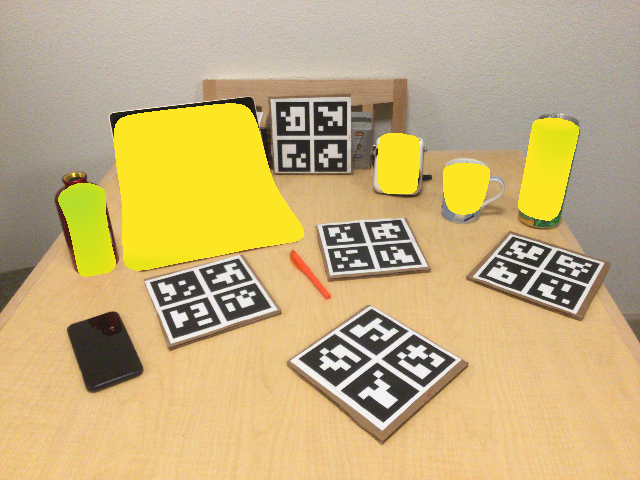}
    \end{subfigure}
    \begin{subfigure}{0.18\linewidth}
        \includegraphics[width=1.0\columnwidth]{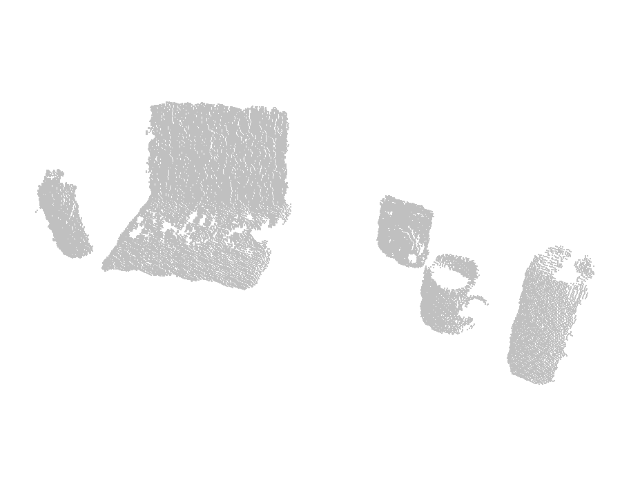}
    \end{subfigure}
    \begin{subfigure}{0.18\linewidth}
        \includegraphics[width=1.0\columnwidth]{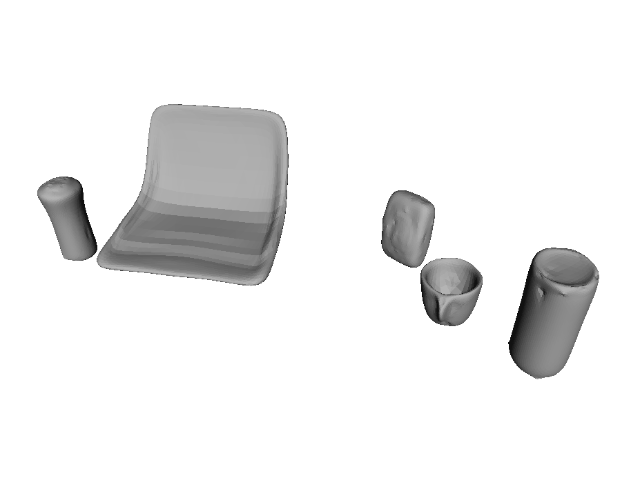}
    \end{subfigure}
    \par\smallskip
    \begin{subfigure}{0.18\linewidth}
        \includegraphics[width=1.0\columnwidth]{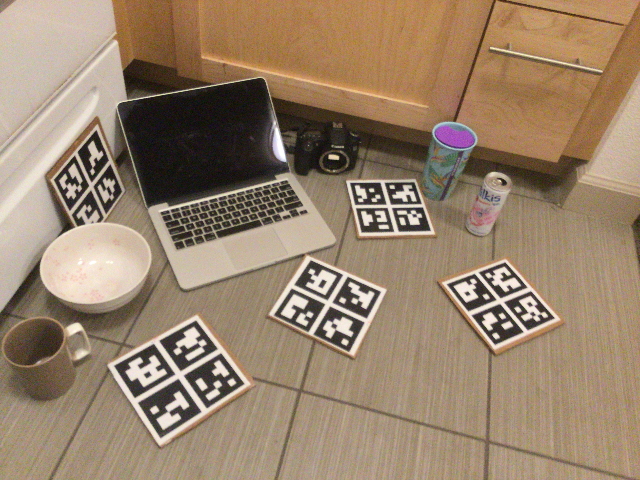}
    \end{subfigure}
    \begin{subfigure}{0.18\linewidth}
        \includegraphics[width=1.0\columnwidth]{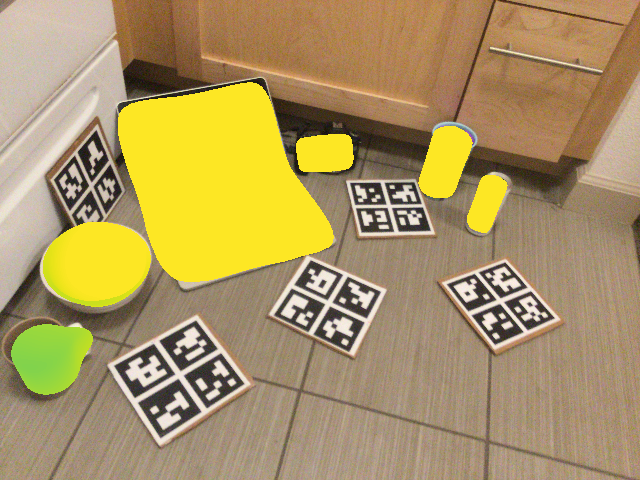}
    \end{subfigure}
    \begin{subfigure}{0.18\linewidth}
        \includegraphics[width=1.0\columnwidth]{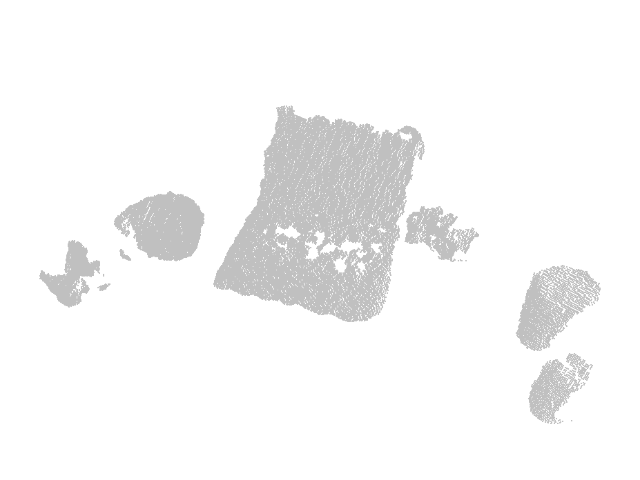}
    \end{subfigure}
    \begin{subfigure}{0.18\linewidth}
        \includegraphics[width=1.0\columnwidth]{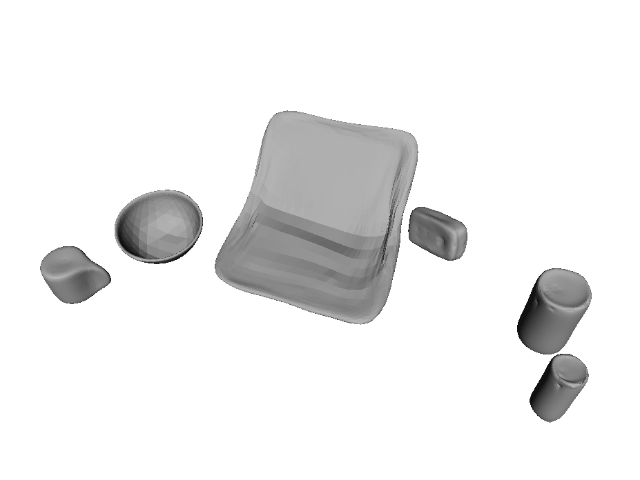}
    \end{subfigure}
    \par\smallskip
    \begin{subfigure}{0.18\linewidth}
        \includegraphics[width=1.0\columnwidth]{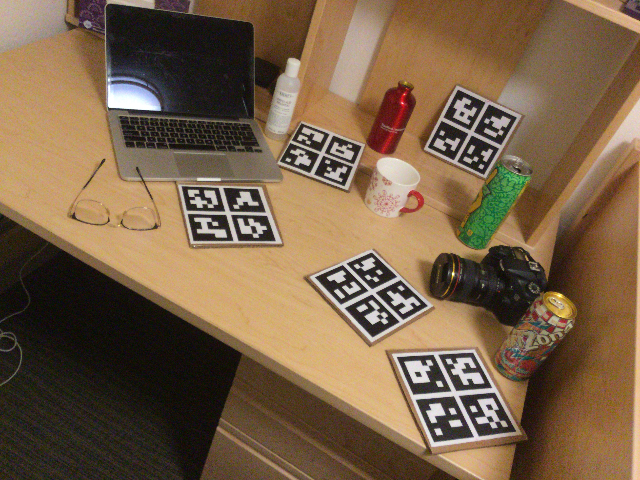}
    \end{subfigure}
    \begin{subfigure}{0.18\linewidth}
        \includegraphics[width=1.0\columnwidth]{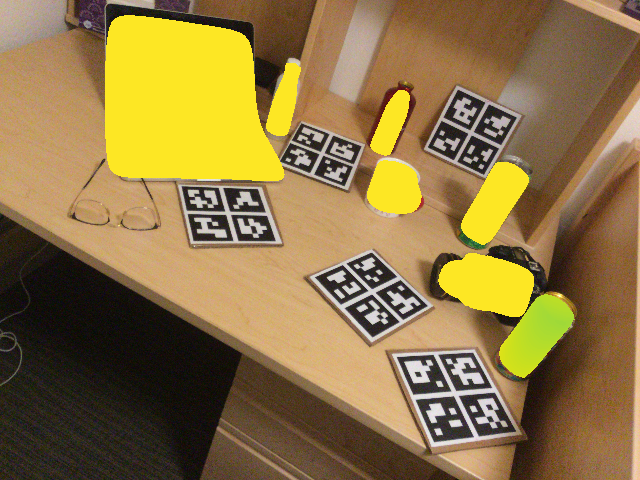}
    \end{subfigure}
    \begin{subfigure}{0.18\linewidth}
        \includegraphics[width=1.0\columnwidth]{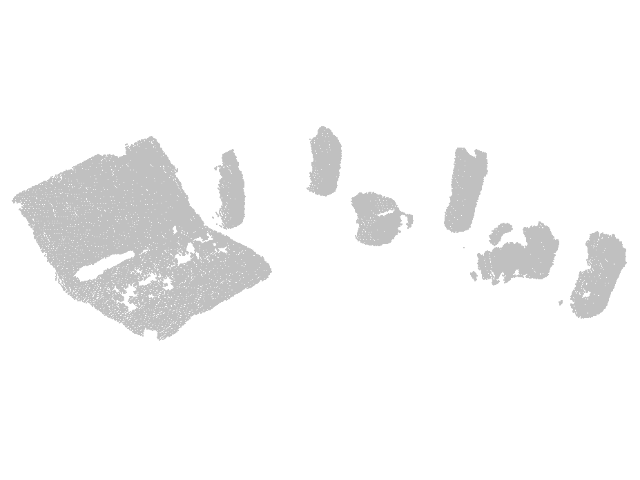}
    \end{subfigure}
    \begin{subfigure}{0.18\linewidth}
        \includegraphics[width=1.0\columnwidth]{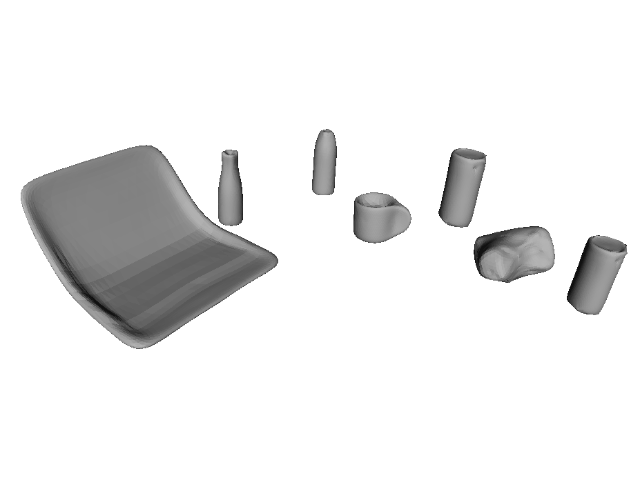}
    \end{subfigure}
\caption{Visualization of example NOCS REAL \cite{wang2019normalized} frames. From left to right are the color image, the color image overlaid by the depth projection of the estimated objects, the back-projected measured depth image, and the estimated objects from a novel viewpoint.} 
\vspace{-1.5em}
\label{fig: nocsviz}
\end{figure*}

%% file: includes/ablation.tex
\begin{table}
\footnotesize
\begin{center}
\caption{Ablation results on NOCS REAL \cite{wang2019normalized}.}
\label{tab: ablation}
{\footnotesize
    \begin{tabular}{|l|c c c c c|}
        \hline
        Ablated Module & IoU75 & 5\degree 2cm & 5\degree 5cm & 10\degree 2cm & 10\degree 5cm \\
        \hline
        \hline
        EM          & -1.5 & -2.1 & -4.0 & -0.4 & +0.2 \\
        \hline
        Symm. Check & -0.3 & -0.6 & -0.5 & -0.4 & -0.4 \\
        \hline
        Rendering   & -0.7 & -2.5 & -2.2 & -0.5 & +0.2 \\
        \hline
        Multi-Hypotheses & -9.9 & -4.9 & -6.2 & -7.8 & -9.3 \\
        \hline
    \end{tabular}
}
\end{center}
\end{table}

%% file: includes/params.tex
\begin{table*}
\begin{center}
\begin{threeparttable}[b]
\caption{Algorithm hyper-parameters.}
\label{tab: params}
    \begin{tabular}{|l|c|}
        \hline
        Description & Value \\
        \hline
        Number of neighbors in depth outlier removal & 500 \\
        \hline
        Number of standard deviations in depth outlier removal & 1 \\
        \hline
        Number of points threshold to ignore a detection & 100 \\
        \hline
        Number of vertices in the mesh template & 2562 \\
        \hline
        Number of points sampled in template deformation (Section \ref{sec: asm} $N = M$) & 5000 \\
        \hline
        $\lambda_n$ & 0.01 \\
        \hline
        $\lambda_e$ & 1.0, 3.0 (mug), 5.0 (camera) \\
        \hline
        $\lambda_l$ & 0.1, 0.01 (mug), 0.3 (can) \\
        \hline
        SGD learning rate & 1.0 \\
        \hline
        SGD momentum & 0.9 \\
        \hline
        Number of PCA components\tnote{1} & 8, 50, 40, 5, 32, 7 \\
        \hline
        Number of pose steps per iteration & 1 \\
        \hline
        Number of shape gradient descent steps per iteration & 5 \\
        \hline
        Total number of iterations where the shape steps are run & 50 \\
        \hline
        Total number of iterations where the pose steps are run & 80 \\
        \hline
        Number of points sampled on the mesh during estimation & 1000 \\
        \hline
        EM $Q$ & 5 \\
        \hline
        EM $\sigma^2$ & 0.2 \\
        \hline
        Mirror Symmetry Objects & all six NOCS objects\tnote{2} \\
        \hline
        Rotation Symmetry Objects & bottle, bowl, mug\tnote{3}, can \\
        \hline
        Number of discrete rotation symmetry & 6 \\
        \hline
        Depth rendering start iteration\tnote{4} & 5 \\
        \hline
        $\lambda_\Psi$ & 1.0 \\
        \hline
        $\lambda_{dr}$ & 1.0 \\
        \hline
        Depth rendering background value (units follow NOCS depth images) & $5\times10^3$ \\
        \hline
        Down-selection iteration & 1, 5, 15 \\
        \hline
        Number of hypotheses kept at Iteration Number & 2304 (initially), 45 at 1, 15 at 5, 1 at 15 \\
        \hline
        Down-selection rotation spacing $\theta$ & 20\degree \\
        \hline
    \end{tabular}
    \begin{tablenotes}
        \item [1] in the order of bottle, bowl, camera, can, laptop, and mug.
        \item [2] Camera is treated as mirror symmetric.
        \item [3] Mug is treated as rotation symmetric since the bulk of it is.
        \item [4] to reduce computation.
    \end{tablenotes}
\end{threeparttable}
\end{center}
\end{table*}

%% file: includes/nocsbenchmark_shape.tex
\begin{table*}
\begin{center}
\caption{Detailed shape estimation results on NOCS REAL \cite{wang2019normalized} evaluated with the Chamfer distance ($10^{-3}$). Baseline results are from sources \cite{irshad2022shapo, akizuki2021asm}. S.D. stands for shape-annotated data required for training and P.D. stands for pose-annotated data required for training. S.C. stands for the shape classification network. The results are roughly in the order of overall performance.}
\label{tab: appnocsshape}
    \begin{tabular}{|l| c c | c c c c c c c|}
        \hline
        Method & P.D. & S.D. & Bottle & Bowl & Camera & Can & Laptop & Mug & Mean \\
        \hline
        \hline
        Categorical Prior \cite{tian2020shape} & \hfill & \hfill & 5.0 & 1.2 & 9.9 & 2.4 & 7.1 & 0.97 & 4.4 \\
        \hline
        \bf{\algName (Ours) w/o S.C.} & \hfill & \hfill & 2.5 & 0.86 & 9.0 & 3.7 & 3.6 & 2.8 & 3.8 \\
        \hline
        ShapePrior \cite{tian2020shape} & \checkmark & \checkmark & 3.4 & 1.2 & 8.9 & 1.5 & 2.9 & 1.0 & 3.2 \\
        \hline
        \bf{\algName (Ours) w/ S.C.} & \hfill & \checkmark & 2.9 & 0.58 & 6.8 & 1.1 & 1.2 & 3.0 & 2.6 \\      
        \hline
        CenterSnap \cite{irshad2022centersnap} & \checkmark & \checkmark & 1.3 & 1.0 & 4.3 & 0.9 & 0.7 & 0.6 & 1.5 \\
        \hline
        ASM-Net \cite{akizuki2021asm} & \checkmark & \hfill & 0.23 & 0.06 & 0.61 & 0.15 & 0.60 & 0.10 & 0.29 \\
        \hline
    \end{tabular}
\end{center}
\end{table*}

%% file: includes/runtime.tex
\begin{table*}
\begin{center}
\caption{Runtime ablation results on NOCS REAL \cite{wang2019normalized}. The case of a single hypothesis, compared to the case of a single initial guess, includes additional one-time upfront computation to select the best hypothesis using $\mathcal{S}_r + \mathcal{S}_\sigma$ at the very beginning of the iteration.}
\label{tab: runtime}
    \begin{tabular}{|l|c|}
        \hline
        Configuration & Average Runtime per Instance (s) \\
        \hline
        \hline
        Full System & 8.0 \\
        \hline
        Given a Single Initial Guess & 5.3 (66\%) \\
        \hline
        Single Hypothesis & 5.7 (70\%) \\
        \hline
        Multi-Hypothesis Pose Estimation Only & 1.9 (23\%) \\
        \hline
    \end{tabular}
\end{center}
\end{table*}

\begin{table*}
\begin{center}
\caption{Runtime breakdown for the most time-consuming processes in the shape estimation step on NOCS REAL \cite{wang2019normalized} given a single initial guess. Pre-processing the measured point cloud includes centering and diagonally normalizing the point cloud. Pre-processing the shape model includes transforming it by the estimated pose, applying the same centering offset and normalization factor as done on the measured point cloud, and meshing the model.}
\label{tab: shaperuntime}
    \begin{tabular}{|l|c|}
        \hline
        Process & Average Runtime per Instance (s) \\
        \hline
        \hline
        Shape Estimation Step Total & 4.6 (88\% in 5.3) \\
        \hline
        Pre-processing the Measured Point Cloud and Shape Model & 0.28 (6\%) \\
        \hline
        Sampling Points on the Mesh Model & 0.62 (13\%) \\
        \hline
        EM Weight Computation for Chamfer & 0.11 (2\%) \\
        \hline
        Normal Consistency Loss \eqref{eq: normal} & 0.49 (11\%) \\
        \hline
        Edge Length Loss \eqref{eq: edge} & 0.57 (12\%) \\
        \hline
        Laplacian Smoothing Loss \eqref{eq: lap} & 0.53 (11\%) \\
        \hline
        Back-Propagation & 1.9 (40\%) \\
        \hline
    \end{tabular}
\end{center}
\end{table*}

%% file: includes/shapeinitabl.tex
\begin{table*}
\begin{center}
\caption{The effects of different shape initialization approaches on NOCS REAL \cite{wang2019normalized} with ground-truth object masks. The results are roughly in the order of overall performance. The best result in each metric is in bold.}
\label{tab: shapeinitabl} 
    \begin{tabular}{|l| c c c c c c c | c|}
        \hline
        Method & IoU25 & IoU50 & IoU75 & 5\degree 5cm & 5\degree 10cm & 10\degree 5cm & 10\degree 10cm & Chamfer($10^{-3}$) \\
        \hline
        \hline
        Mean Shape & \bf{97.1} & 65.8 & 40.3 & 26.1 & 26.4 & 38.2 & 38.7 & 4.05 \\
        \hline
        Mean Code & \bf{97.1} & 66.0 & 40.3 & 26.0 & 26.3 & 38.2 & 38.8 & 4.05 \\
        \hline
        CNN Closest & 96.2 & \bf{67.5} & 49.6 & 36.7 & 36.9 & 48.6 & 49.1 & 2.39 \\
        \hline
        GT Closest & 97.0 & 62.6 & \bf{51.0} & \bf{38.9} & \bf{39.1} & \bf{50.5} & \bf{51.0} & \bf{1.18} \\
        \hline
    \end{tabular}
\end{center}
\end{table*}

%% file: includes/shapeeffectsonpose.tex
\begin{table*}
\begin{center}
\caption{The effects of shape accuracy (reported in terms of the Chamfer distance) on pose estimation on NOCS REAL \cite{wang2019normalized} with ground-truth object masks. The results are roughly in the order of overall performance. The best result in each metric is in bold.}
\label{tab: shapeeffpose} 
    \begin{tabular}{|l| c | c c c c c c c|}
        \hline
        Given Shape & Chamfer($10^{-3}$) & IoU25 & IoU50 & IoU75 & 5\degree 5cm & 5\degree 10cm & 10\degree 5cm & 10\degree 10cm \\
        \hline
        \hline
        Mean Shape       & 4.67 & \bf{97.1} & 65.8 & 35.5 & 24.9 & 25.6 & 37.4 & 39.0 \\
        \hline
        CNN Closest ShapeNet & 2.27 & 94.5 & 67.6 & 48.3 & 29.9 & 30.1 & 46.1 & 46.7 \\
        \hline
        CNN Closest Template & 2.47 & 95.9 & \bf{68.1} & 48.6 & 30.4 & 30.7 & 47.6 & 48.2 \\
        \hline
        CNN Closest PCA      & 2.51 & 95.9 & 67.7 & 48.8 & 31.9 & 32.2 & 46.7 & 47.3 \\
        \hline
        GT Shape         & 0.15 & 95.0 & 63.9 & 50.9 & 36.6 & 36.9 & 48.1 & 48.7 \\
        \hline
        GT Closest ShapeNet & 0.75 & 96.7 & 65.7 & \bf{52.6} & 37.1 & 37.3 & 50.8 & 51.3 \\
        \hline
        GT Closest Template & 0.88 & 96.6 & 63.9 & 51.6 & 37.8 & 37.9 & 50.9 & 51.3 \\
        \hline
        GT Closest PCA      & 0.93 & 96.9 & 62.4 & 51.8 & \bf{37.9} & \bf{38.0} & \bf{51.0} & \bf{51.4} \\
        \hline
    \end{tabular}
\end{center}
\end{table*}

%% file: includes/sensitivity_multihyp.tex
\begin{table*}
\begin{center}
\caption{\diffcolor{The effects of different numbers of hypotheses and different iteration numbers to down-select hypotheses. "Nmb at Iter" means the number of hypotheses kept at the iteration numbers where the hypotheses undergo down-selection. The first row is the default setting}}
\label{tab: sens_multihyp}
    \begin{tabular}{|l| c | c c c c c c c | c |}
        \hline
        Nmb at Iter & Runtime (s) & IoU25 & IoU50 & IoU75 & 5\degree 5cm & 5\degree 10cm & 10\degree 5cm & 10\degree 10cm & Chamfer($10^{-3}$) \\
        \hline
        \hline
        45 15 1 at 1 ~5 15     & 8.0 & 82.1 & 58.0 & 42.2 & 36.5 & 37.1 & 50.6 & 51.4 & 2.59 \\
        \hline
        90 30 1 at 1 ~5 15     & 9.5 (120\%) & +0.5 & +0.1 & -0.2 & +0.4 & +0.5 & -0.1 &  0.0 & -0.05 \\
        \hline
        23 ~8 1 at 1 ~5 15    & 6.7 (84\%)  & -1.7 & -1.2 & -1.7 & -1.6 & -1.7 & -2.6 & -2.6 & +0.09 \\
        \hline
        45 15 1 at 1 10 30      & 9.5 (119\%)  & 0.0  & -1.2 & -1.4 & -1.1 & -1.1 & -1.9 & -1.9 & +0.03 \\ 
        \hline
        45 15 1 at 1 ~3 ~7    & 6.7 (83\%) & +0.7 & +0.6 & +0.2 & -1.9 & -2.0 & +1.3 & +1.3 & -0.01 \\
        \hline
    \end{tabular}
\end{center}
\end{table*}

%% file: includes/debugged.tex
\begin{table*}
\begin{center}
\caption{Updated results on NOCS REAL \cite{wang2019normalized} with debugged NOCS evaluation code.}
\label{tab: debugupdate}
    \begin{tabular}{|l| c c c c c c c|}
        \hline
        Method & IoU25 & IoU50 & IoU75 & 5\degree 5cm & 5\degree 10cm & 10\degree 5cm & 10\degree 10cm \\
        \hline
        \hline
        \multicolumn{8}{| c |}{Updated Table \ref{tab: nocspose}} \\
        \hline
        \hline
        \algName~(Ours) w/o S.C. & 82.0 & 64.2 & 25.7 & 32.5 & 33.5 & 42.4 & 43.8 \\
        \hline
        \algName~(Ours) w/ S.C.  & 80.6 & 60.9 & 24.1 & 36.7 & 37.3 & 50.8 & 51.7 \\
        \hline
        \hline
        \multicolumn{8}{| c |}{Updated Table \ref{tab: shapeinitabl}} \\
        \hline
        \hline
        Mean Shape  & 97.1 & 77.3 & 31.4 & 26.3 & 26.5 & 38.4 & 39.0 \\
        \hline
        Mean Code   & 97.0 & 77.4 & 31.5 & 26.2 & 26.4 & 38.5 & 39.1 \\
        \hline
        CNN Closest & 96.6 & 76.5 & 31.5 & 36.9 & 37.2 & 48.8 & 49.4 \\
        \hline
        GT Closest  & \bf{97.2} & \bf{83.4} & \bf{37.3} & \bf{39.1} & \bf{39.2} & \bf{50.8} & \bf{51.3} \\
        \hline
        \hline
        \multicolumn{8}{| c |}{Updated Table \ref{tab: shapeeffpose}} \\
        \hline
        \hline
        Mean Shape       & \bf{96.8} & 75.5 & 26.2 & 25.1 & 25.8 & 37.7 & 39.2 \\
        \hline
        CNN Closest ShapeNet & 95.7 & 72.2 & 25.9 & 29.9 & 30.1 & 46.1 & 46.7 \\
        \hline
        CNN Closest Template & 95.4 & 69.7 & 31.4 & 30.5 & 30.8 & 47.7 & 48.3 \\
        \hline
        CNN Closest PCA      & 95.5 & 71.9 & 29.8 & 32.1 & 32.3 & 46.9 & 47.4 \\
        \hline
        GT Shape         & 95.8 & \bf{83.4} & \bf{39.5} & 36.3 & 36.7 & 47.9 & 48.5 \\
        \hline
        Closest ShapeNet & 95.9 & 76.8 & 34.2 & 37.2 & 37.4 & 51.0 & \bf{51.5} \\
        \hline
        Closest Template & 96.7 & 79.5 & 35.1 & \bf{37.8} & \bf{38.0} & \bf{51.1} & \bf{51.5} \\
        \hline
        Closest PCA      & 96.5 & 79.7 & 35.6 & \bf{37.8} & \bf{38.0} & 51.0 & 51.4 \\
        \hline
        \hline
        \multicolumn{8}{| c |}{Updated Table \ref{tab: sens_multihyp}} \\
        \hline
        \hline
        45 15 1 at 1 ~5 15    & 80.6 & 60.9 & 24.1 & 36.7 & 37.3 & 50.8 & 51.7 \\
        \hline
        90 30 1 at 1 ~5 15    & -0.1 & +1.5 & +0.7 & +0.3 & +0.4 & -0.2 & -0.1 \\
        \hline
        23 ~8 1 at 1 ~5 15    & +0.2 & +0.7 & -2.0 & -1.6 & -1.7 & -2.6 & -2.6 \\
        \hline
        45 15 1 at 1 10 30    & +0.2 & +1.0 & 0.0  & -1.2 & -1.2 & -1.9 & -1.9 \\ 
        \hline
        45 15 1 at 1 ~3 ~7    & 0.3  & +2.3 & +0.1 & -2.0 & -2.0 & +1.3 & +1.3 \\
        \hline
    \end{tabular}
\end{center}
\end{table*}

\begin{table*}
\begin{center}
\caption{Updated Table \ref{tab: ablation} with debugged NOCS evaluation code.}
\label{tab: debugablation}
    \begin{tabular}{|l|c c c c c|}
        \hline
        Ablated Module & IoU75 & 5\degree 2cm & 5\degree 5cm & 10\degree 2cm & 10\degree 5cm \\
        \hline
        \hline
        EM          & +3.2 & -2.1 & -4.0 & -0.4 & +0.2 \\
        \hline
        Symm. Check & -0.1 & -0.6 & -0.6 & -0.5 & -0.4 \\
        \hline
        Rendering   & -0.2 & -2.5 & -2.3 & -0.6 & +0.2 \\
        \hline
        Multi-Hypotheses & -2.1 & -4.9 & -6.3 & -8.0 & -9.5 \\
        \hline
    \end{tabular}
\end{center}
\end{table*}